%% file: templateArxiv.tex
\documentclass{article}

\usepackage{PRIMEarxiv}

\usepackage[utf8]{inputenc} 
\usepackage[T1]{fontenc}    
\usepackage{hyperref}       
\usepackage{url}            
\usepackage{booktabs}       
\usepackage{amsfonts}       
\usepackage{nicefrac}       
\usepackage{microtype}      
\usepackage{lipsum}
\usepackage{fancyhdr}       
\usepackage{graphicx}       
\usepackage{subcaption}
\graphicspath{{media/}}     
\usepackage{geometry}
\usepackage{xcolor}
\usepackage{listings}
\definecolor{codegray}{rgb}{0.5,0.5,0.5}
\definecolor{codegreen}{rgb}{0,0.5,0}
\definecolor{codepurple}{rgb}{0.58,0,0.82}
\definecolor{backcolour}{rgb}{0.96,0.96,0.96}
\lstdefinestyle{terminal}{
  basicstyle=\ttfamily\small,
  backgroundcolor=\color{backcolour},
  breakatwhitespace=false,
  breaklines=true,
  keepspaces=true,
  showspaces=false,
  showstringspaces=false,
  showtabs=false,
  tabsize=2,
  frame=single,
  rulecolor=\color{codegray},
  columns=fullflexible
}
\lstdefinestyle{jsonstyle}{
  basicstyle=\ttfamily\small,
  backgroundcolor=\color{backcolour},
  breaklines=true,
  keepspaces=true,
  showstringspaces=false,
  numbers=left,
  numberstyle=\tiny\color{codegray},
  numbersep=8pt,
  tabsize=2,
  frame=single,
  rulecolor=\color{codegray},
  stringstyle=\color{codepurple},
  commentstyle=\color{codegreen},
  columns=fullflexible
}
\lstdefinelanguage{toml}{
  comment=[l]{\#},
  morecomment=[s]{[}{]},
  morekeywords={true,false},
  morestring=[b]{"},
  morestring=[b]{'},
  basicstyle=\ttfamily\small,
  showstringspaces=false,
  breaklines=true,
  columns=fullflexible
}
\usepackage{titlesec}
\usepackage{parskip}
\usepackage{hyperref}
\usepackage{mdframed}
\usepackage{fontawesome5}  
\usepackage{lmodern}
\usepackage{enumitem}
\usepackage[most]{tcolorbox}
\tcbuselibrary{breakable,skins}
\usepackage{float}
\usepackage[utf8]{inputenc}
\usepackage{longtable}
\usepackage{array}
\usepackage[table]{xcolor}
\usepackage{booktabs}

\newcounter{stepcounter}

\title{AI Coding Agents in Social Science: Methodologically Diverse, Empirically Consistent, Interpretively Vulnerable
}

\author{
  Meysam Alizadeh\\
  University of Oxford\\
  University of Zurich\\
  \And
  Fabrizio Gilardi \\
  University of Zurich \\
   \And
  Mohsen Mosleh \\
  University of Oxford \\
   \AND
  Enkelejda Kasneci \\
  Technical University of Munich \\
}

\begin{document}
\maketitle

\subsection*{Significance}
Concerns about LLMs in science pull in two directions: AI homogenization, where agents compress methodological diversity, and credibility-revolution concerns that analytic flexibility enables motivated reasoning. We argue addressing that these concerns require distinguishing between design and verdict layers. Using a many-analysts study, we evaluate two frontier coding agents under neutral and biased prompting. The agents rival human methodological diversity while producing effect estimates close to them. A prompt-induced researcher prior reshuffles methodological decisions without shifting estimates and verdicts. By contrast, an explicit instruction to seek hypothesis-supporting findings flips one agent’s verdicts from 10\% to 90\% support while leaving its coefficient distribution essentially unchanged. The main risk of AI-assisted science may therefore not be homogenized analysis, but insufficiently constrained interpretation.

\vspace{5mm}

\begin{abstract}

The deployment of LLM-based agents in scientific analysis raises opposing concerns: that agents may reduce methodological diversity, or that they may amplify the analytic flexibility through which researchers reach motivated conclusions. We argue these worries target two empirically separable layers: a \emph{design layer} of methodological choices, and a \emph{verdict layer} in which a decision rule maps estimates to a substantive claim. We test both by running 20 independent executions of Claude Code and Codex on a prominent immigration and social-policy against a many-analysts human baseline. At the design layer, Codex matches human methodological diversity and Claude Code produces nearly three times as many specifications; both agents' effect estimates remain broadly aligned with the human consensus, and no agent model exactly matches any human model. A prompt-induced anti-immigration researcher prior reorganizes each agent's methodological decisions but, unlike for biased human analysts in the same data, does not shift aggregate estimates or final verdicts; nor do agents reroute along the methodological axes humans use to bias their estimates. At the verdict layer, an explicit confirmatory prompt flips Claude Code's verdicts from 10\% to 90\% support while leaving its coefficient distribution essentially unchanged, operating through rule omission rather than rule softening. AI agents can rival or exceed human methodological diversity at the design layer while remaining vulnerable at the verdict layer. In our setting, the locus of AI bias is not estimation but interpretation.

\end{abstract}


\section{Introduction}

Scientific discovery depends not only on the availability of data, but also on the diversity of methods used to interpret it \cite{longino2002fate, mitchell2003biological}. Across disciplines, progress has historically emerged from methodological pluralism, in which competing analytical strategies generate alternative explanations tested against empirical evidence \cite{hong2004groups, devezer2019scientific}, collectively shaping scientific understanding \cite{liu2023data}. Such diversity is particularly important in research on human societies, where core concepts and quantities are often open to broad interpretation \cite{breznau2022observing}.

Yet the same methodological pluralism that can support discovery is also a substrate on which uncertainty, bias, and opportunism can act. In the many-analysts study that anchors our benchmark, 73 independent teams analyzing identical data reached effect estimates ranging from strongly negative to strongly positive \cite{breznau2022observing}, and a re-analysis of those data showed that researchers' prior views was associated with their model specifications and reported conclusions \cite{borjas2026ideological}. The same researcher degrees of freedom that enable productive exploration also can enable selective reporting and $p$-hacking through ``garden of forking paths'' decisions \cite{simmons2011false, gelman2013garden}. Methodological diversity is therefore Janus-faced: a driver of collective discovery when transparent and aggregated across the field, and a vehicle for motivated inference when concentrated within a single analysis.

Recent advances in LLM-based agents increasingly support automated execution of substantial parts of the research workflow, including code generation, replication of published analyses, and machine-learning experimentation \cite{lu2024ai, jimenez2024swe, starace2025paperbench, starace2025paperbench}. As these systems increasingly participate in methodological decision-making, both faces of diversity become acute. LLMs often show reduced creative diversity in problems without definitive ground-truth answers \cite{li2024predicting, west2025base, zhang2025noveltybench, zhang2025forcing}, raising concerns about AI homogenization \cite{bommasani2022picking, jiang2026artificial, sourati2025shrinking}. At the same time, LLMs exhibit sycophancy toward user framings \cite{sharma2024towards, perez2023discovering} and susceptibility to reward- and specification-hacking \cite{pan2022effects, denison2024sycophancy}, raising the symmetric concern that whatever diversity they do produce may be steerable by prompt framing. Observational social science is a useful test case for both concerns: core constructs such as socioeconomic status or partisanship are inherently unobservable and admit multiple competing operationalizations \cite{bramson2017understanding, jacobs2021measurement, benoit2019measuring}, which reflect broader theoretical and normative assumptions about what a construct should capture \cite{shadish2002experimental, kusmaryono2022number}, leaving substantial room for both convergence and motivated divergence.

The two concerns (too little diversity, or diversity of the wrong kind) are typically discussed as if they were one. We argue they are not, and that pulling them apart is the conceptual move useful to evaluate AI agents in scientific workflows. We analyze agent behavior at two layers. The \emph{design layer} consists of methodological choices about measurement, sample definition, model specification, estimator selection, uncertainty quantification, and robustness checks. The \emph{verdict layer} consists of mapping empirical estimates onto a substantive verdict about the hypothesis (e.g.\ concluding that a hypothesis is supported if four of six estimates are negative and statistically significant at $p < 0.05$), and the faithful narration of the decision rule's output. Diversity is epistemically productive at the design layer, because more methodological pathways means more of the multiverse is probed. But at the verdict layer, discipline is epistemically essential. Without it, design-layer diversity becomes a menu of conclusions to choose from. The two layers are conceptually independent: an agent can be high-diversity and high-discipline (exhaustive exploration with a pre-committed mapping to conclusions), or high-diversity and low-discipline (exhaustive exploration with the conclusion chosen after the fact). Across 73 human research teams, Breznau et al. \cite{breznau2022observing} found that the share of statistically supportive test results explained only about a third of the deviance in narrated conclusions, suggesting that estimates and verdicts can come apart even when no one team is selecting between them. For AI coding agents, where prompt interventions can target each layer separately, the same distinction can be measured rather than inferred. 

Building on the many-analysts dataset of \cite{breznau2022observing}, in which 73 research teams independently tested whether greater immigration reduces public support for social policy \cite{brady2014does} using identical data, we evaluate twenty independent runs of two frontier coding agents (Claude Code and Codex) on the same task. Our experiments yield three key findings. First, frontier coding agents do not, in our setting, collapse toward a single canonical analytic strategy: Codex matches the methodological diversity of human analyst teams, and Claude Code substantially exceeds it, while both produce effect distributions and substantive conclusions broadly consistent with the human baseline. This complicates homogenization narratives at the design layer. Second, unlike the pattern reported for human researchers in this benchmark, a prompt-induced researcher prior reshuffles each agent's methodological pathways without shifting aggregate estimates or final verdicts. Importantly, agents do not shift along the methodological choices through which anti-immigration human researchers shift their estimates in the same data \cite{borjas2026ideological}. Third, the design and verdict layers are empirically separable in agent-led analysis: a confirmatory prompt that instructs the agent to select hypothesis-supporting results leaves Claude Code's coefficient distribution essentially unchanged while flipping its verdicts from 10\% support to 90\%, and a prompt-injected researcher prior shifts methodological pathways without shifting aggregate estimates or final verdicts in either agent. The locus of prompt-induced bias is therefore not estimation but narration, which is a failure mode that would be missed by evaluations that summarize agents only by their numerical outputs.


\section{Results}
Before presenting the results, we briefly summarize the experimental setup (see Materials \& Methods for full details). Each agent, Claude Code (Opus 4.7 1M, ``Max Effort'') and Codex (GPT~5.5, ``Extra High Intelligence''), completed twenty independent runs of the same task: testing the hypothesis that higher immigration reduces public support for social policy, using the original International Social Survey Programme (ISSP) data and country-level macroeconomic indicators. Both agents received the identical natural-language prompt; no agent specific wording, hints, or scaffolding were used. Each agent operated within a sandboxed working directory that confined file-system access to the provided replication materials, but within that sandbox the agents were permitted to install Python and R packages and to perform unrestricted web searches, mirroring the resources available to the human research teams in the original crowdsourced study. Each run encompassed the full pipeline including research design, code authorship, execution, and written conclusion, and proceeded in fully automated mode, with no human intervention during agent execution and no memory of any prior run.

\begin{figure}[t]
  \centering
  \includegraphics[width=0.7\linewidth]{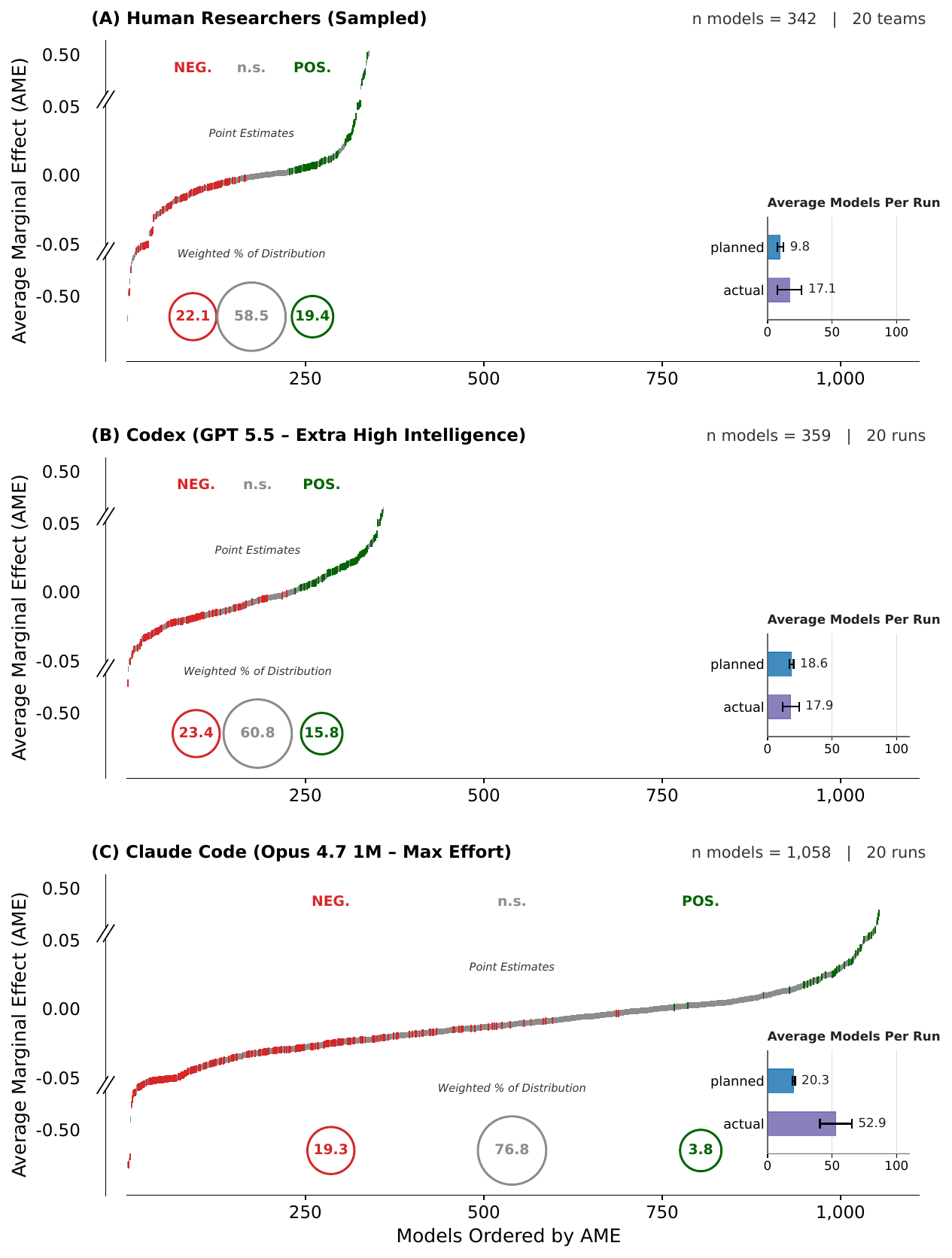}
  \caption{\textbf{Specification curves of standardized average marginal effects (AMEs)} for the hypothesis that immigration reduces public support for social policy. Each hash mark is one converged model, ordered along the x-axis by AME within each panel; color denotes the 95\% CI relative to zero ({\color[HTML]{D62728}red}: negative-significant; grey: includes zero; {\color[HTML]{006400}dark green}: positive-significant). The y-axis is piecewise-compressed with breaks at $\pm 0.05$. Circles report team/run-weighted percentages of models in each category (weights $1/n_{\text{models per team/run}}$ so each team/run contributes equally). The lower-right inset contrasts the number of models implied by each team's pre-registered factor grid (\emph{planned}) with the number actually executed (\emph{actual}); bars are means and error bars are 95\% CIs. (\textit{A})~Twenty teams drawn at random (seed $=42$) from the 73 teams of ref.~\cite{breznau2022observing}; $n = 342$ models. (\textit{B})~Claude Code (Opus~4.7~1M, ``Max Effort''),
    20 runs; $n = 1{,}058$. (\textit{C})~Codex (GPT~5.5, ``Extra High Intelligence''), 20 runs; $n = 359$.}
  \label{fig:spec_curves}
\end{figure}

\subsection{Comparing AI Agents and Human Researchers in
Methodological Diversity}

Fig.~\ref{fig:spec_curves} replicates the headline visualization of the original crowdsourced replication initiative (CRI) \cite{breznau2022observing}, a per-team rendering of the standardized average marginal effect (AME) distribution, and extends it to two frontier coding agents. We compare three matched-size groups: a random sample of 20 of the 73 human research teams (panel \textit{A}; seed $=42$), 20 independent runs of Claude Code (panel \textit{B}), and 20 independent runs of Codex (panel \textit{C}). Each hash mark is one converged model; models within a panel are ordered along the x-axis by AME, and the three panels share a common x-axis range (0--1{,}100) so that the horizontal extent of each hash-mark block is proportional to the total number of executed specifications.

\paragraph{Specification effort differs sharply between agents.}
Across 20 attempts each on the identical task, CC delivered 1{,}058 valid AME estimates (mean $52.9 \pm 26.4$ SD, median 55, IQR 32--71, range 14--107), whereas CX delivered only 359
(mean $17.9 \pm 13.7$, median 16, IQR 14--18, range 2--58). The 20
sampled human teams produced a per-team volume similar to CX
(mean 15.9, range 1--54). The ratio of mean per-run specifications
between CC and CX was 2.95 (bootstrap 95\% CI 2.01--4.36), with the gap robust to non-parametric testing (Mann--Whitney $U = 354$,
$P = 3.3 \times 10^{-5}$; rank-biserial $r = 0.77$) and to a Welch
$t$ test on log counts ($t = 5.05$, $P = 1.6 \times 10^{-5}$).
CX was more variable than CC in relative terms (coefficient of
variation 0.76 vs.\ 0.50), with three runs returning only two
specifications each, suggesting that CX may have terminated after a minimal stock-and-flow analysis, versus a long upper tail in CC that included one run with 107 specifications.

Across all three groups, the modal outcome was a 95\% confidence interval that includes zero ($58.5\%$, $76.8\%$, and $60.8\%$ of models for humans, CC, and CX), preserving the null finding of Brady and Finnigan \cite{brady2014does}. Where the groups diverge most visibly is in the \emph{shape} and \emph{volume} of their executed specification space. The sampled human teams produced an approximately symmetric mix of significant findings ($22.1\%$ negative, $19.4\%$ positive); CX produced a slightly less symmetric but qualitatively similar mix ($23.4\%$, $15.8\%$); whereas CC produced a strongly asymmetric distribution ($19.3\%$ negative, only $3.8\%$ positive), with the bulk of its $1{,}058$ estimates pulled into the non-significant central mass. 

\paragraph{Agents plan bigger; humans stay closer to plan.}
For every team and run we computed the number of model specifications \emph{implied by the pre-registered plan} ($n_{\text{planned}}$) alongside the actual delivered count ($n_{\text{actual}}$). For human teams, $n_{\text{planned}}$ is the implied factor grid of the registered design: distinct dependent variables (any non-zero proportion in \texttt{\{Jobs, Unemp, IncDiff, OldAge, House, Health, Scale\}} of \texttt{cri\_team.csv}) times distinct immigration measures (\texttt{\{Stock, Flow, ChangeFlow\}}). For CC and CX, each run's \texttt{research\_design.md} was parsed for the same six dependent variables (plus composite index) and the three immigration measures, and $n_{\text{planned}}$ is the product. On paper, the three groups committed to comparable ambition: $n_{\text{planned}}$ averaged $9.8$ per human team ($4$--$21$), $20.3$ per CC run ($14$--$21$), and $18.6$ per CX run ($14$--$21$); the $\geq\!12$ threshold (the natural minimum for a six-DV, two-measure design) was met by 40\% of human teams but 100\% of CC and CX runs. Execution diverged in different directions: humans over-delivered (actual/planned $1.8\times$) through undocumented robustness specifications, CC over-delivered by an order of magnitude ($2.6\times$) with every run exceeding its own grid, and CX tracked its plan in the mean ($\approx 1.0\times$) but with an actual spread far wider than its tight planned range. 

\subsection{Comparing AI Agents and Human Researchers in Estimate Similarity}
Volume and method-mix differences tell us how the three groups \emph{search} the analytic space, but say nothing about whether they \emph{arrive at the same answers}. A coverage advantage of agents is only meaningful if the resulting effect estimates remain comparable to those produced by domain experts; agents that explore three times more specifications but settle on a systematically different distribution of AMEs would make the findings appear more robust without actually producing similar conclusions. To test this, we performed two complementary analyses. First, we compared the empirical distribution of per-cell AMEs produced by each agent against the distribution of human-team estimates, separately for each of the seven dependent variables (six item-level outcomes plus the composite social-policy scale), using the two-sample Kolmogorov--Smirnov distance $D$ as a non-parametric distributional test (SI Fig.~\ref{fig:fig5a}). Second, we evaluated whether agents could accurately reproduce the original results reported in Brady and Finnigan (2014), specifically Tables 4 and 5, under five levels of information availability, ranging from only data and contextual access to full access to the original materials. This second test assesses whether agents can accurately recover published target estimates under varying informational constraints.

\paragraph{Agents and humans mostly agree on effect estimates, with one systematic exception.}

The first test compares the full distribution of AMEs produced by agents and humans across outcomes. As shown in SI Fig.~\ref{fig:fig5a}, on the four single-item outcomes that anchor the original debate (jobs, unemployment, income difference, and old age), we do not reject equality of distributions between both agents and the 20-team humans at $\alpha = 0.05$ (CC: $D = 0.22, 0.17, 0.24, 0.18$; CX: $D = 0.19, 0.21, 0.28, 0.31^{}$, with only the old-age comparison for Codex reaching significance). The two agents diverge from humans on distinct subsets of the remaining outcomes: Claude Code's distributions are significantly compressed on housing, health, and the composite scale ($D = 0.21^{}, 0.35^{}, 0.38^{}$), while Codex diverges only on the composite ($D = 0.35^{*}$). The composite scale is thus the single outcome on which both agents depart in our tests from human practice, and the divergence is in the same direction for both, suggesting that the gap is more consistent with a shared tendency to construct the composite from a more uniform subset of items than the heterogeneous, theory-led aggregations used by human teams, than by agent-specific quirks. Taken together, these findings suggest that even though agents often explore substantially larger specification spaces, their resulting estimates generally remain close to the range of human conclusions, with disagreements concentrated around a specific construct rather than reflecting broad miscalibration.

\paragraph{Agents reproduce qualitative conclusions, but exact estimates only when code is provided.}
We assessed whether two LLM-based coding agents could reproduce the 72 country-level coefficients in Tables~4 and 5 of Brady and Finnigan \cite{brady2014does} under five conditions of increasing transparency, from the research question alone to full access to methods and code (SI Fig.~\ref{fig:overview-accuracy}). Both agents converge to perfect reproduction once the original code is supplied (100\% exact match under \textit{Model + Results + Code} and \textit{Full Access}), but below this threshold exact numerical reproduction is essentially unattainable: the joint exact match on the significance marker, odds ratio, and \textit{z}-score (rounded to the paper's 3-decimal precision) stays below $1\%$ on average for Claude Code across all partial-information conditions and is negligible for Codex with methods only (1.1\%), rising to 39.4\% for Codex under \textit{Model + Results} (SI Fig.~\ref{fig:overview-accuracy}\textit{A,B}); the odds ratio alone partially survives, reaching 17.2\% (Claude Code) and 58.1\% (Codex) when the specification is provided (panel~\textit{C}). Qualitative inference is far more robust: requiring only agreement on significance and sign, accuracy reaches 68.6\%/77.8\% (Claude Code/Codex) from methods alone and exceeds 91\% across all model-aware conditions (panel~\textit{D}). Codex outperforms Claude Code in every partial-information condition, with two of five \textit{Model + Results} runs achieving $\geq\!95.8\%$ numerical reproduction while others cluster near zero---plausibly because some runs recover the original Stata estimation routine and rounding convention while others adopt internally consistent but divergent Python implementations.

\paragraph{The residual appears to reflect documentation gaps more than agent capability.}
Inspecting the cells where agents miss the published values, the persistent errors we inspected primarily reflect a documentation gap rather than a limit of the agents, via two mechanisms. First, four sample-construction choices---the handling of a 999{,}996 no-answer sentinel in the 1996 ISSP household-income variable, the imputation of self-employment for respondents outside the labour force, the mapping of wave-specific ISSP education codes (\texttt{v205} in 1996, \texttt{DEGREE} in 2006) onto the paper's three-category taxonomy, and the choice of omitted country in the fixed-effect dummies---appear in neither the main text nor the supplement; the authors' archived Stata do-file resolves all four, but without it agents must infer these choices, and each defensible reading differs across runs, producing analytical-$N$ drift of up to ${\sim}1{,}000$ respondents per dependent variable and the run-to-run variance in SI Fig.~\ref{fig:overview-accuracy}\textit{A--C}. Second, a typographical inconsistency in Table~5---the B3~$\times$~retirement~$\times$~Net Migration cell printed as $1.128^{**}$ despite a $z$-score of $2.458$ that the legend maps to a single star---propagates one significance-marker mismatch into every reproduction, including the otherwise-perfect \textit{Full Access} runs ($1/72 \approx 1.4\%$ of cells; the residual in Fig.~\ref{fig:overview-accuracy}\textit{D}). Neither failure mode appear to be primarily an inferential limit of the agents: the first is information the study did not release, the second is an inconsistency in the published values themselves. The bottleneck separating ``conclusion-accurate'' from ``digit-accurate'' reproduction is therefore the documentation practices of the original study, pinning the remaining variance to fixable gaps in scholarly disclosure.

\subsection{Profiling method choices across humans, Codex, and Claude Code}

\begin{figure*}[t]
  \centering
  \includegraphics[width=\textwidth]{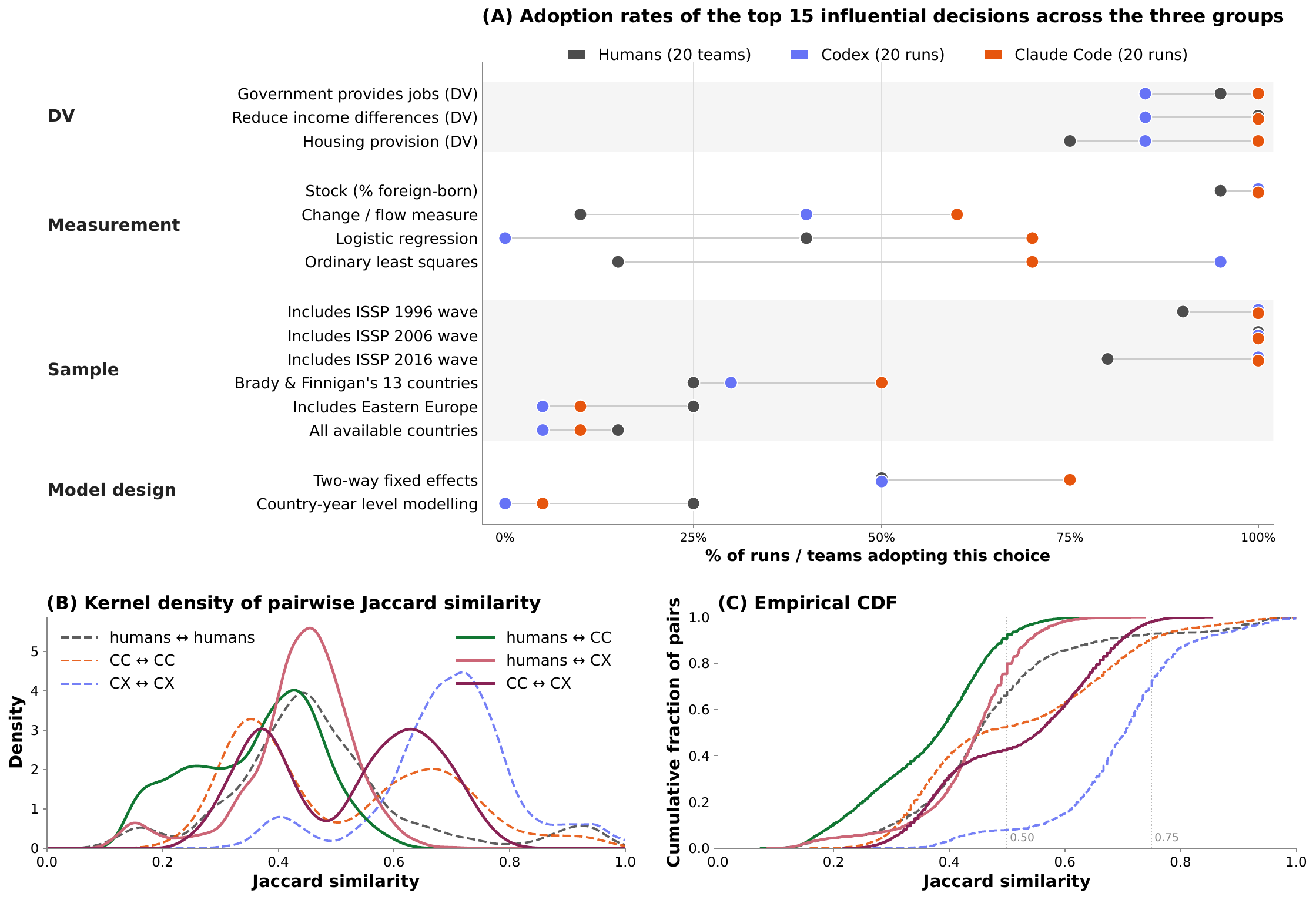}
  \caption{\textbf{Method-choice comparison across human researchers and AI agents.}
Per-model decision flags were extracted from each unit's executed
models: parsed marginal-effects output and replication code for
20 Codex (CX) and 20 Claude Code (CC) runs, and the matching flag
columns in \texttt{cri.csv} for 20 human teams drawn at random
(seed $=42$) from \cite{breznau2022observing}.
(\textit{A}) Percentage of units (runs/teams) in which at least one
executed model uses each of the 15 most influential decisions of
Breznau et al.---the \emph{m07adj} predictor block from
\emph{04\_CRI\_Main\_Analyses.Rmd}, grouped into four themes (DV, Measurement, Sample, Model design).
(\textit{B}) Kernel-density estimates and (\textit{C}) empirical CDFs
of pairwise Jaccard similarity between every two executed models,
computed over the 174 substantive S12 decisions (193 S12 variables
minus 19 administrative identifiers and 3 PI-uncoded country rows).
Each panel shows six distributions: three within-group baselines
(\emph{dashed}: humans$\leftrightarrow$humans, CC$\leftrightarrow$CC,
CX$\leftrightarrow$CX; self-pairs excluded) and three between-group
pairs (\emph{solid}: humans$\leftrightarrow$CC,
humans$\leftrightarrow$CX, CC$\leftrightarrow$CX). Reference lines in
(\textit{C}) mark Jaccard $=0.50$ and $0.75$.}
  \label{fig:method_comparison}
\end{figure*}

To see how methodological decisions vary across humans and the two
AI agents, we compared the three groups along two complementary
axes (Fig.~\ref{fig:method_comparison}). First, we calculated the adoption rate of the 15 most influential methodological decisions identified by Breznau et al. \cite{breznau2022observing} (see \textit{Materials \& Methods}), grouped into dependent-variable choice, measurement, sample construction, and model specification. Following Breznau et al. \cite{breznau2022observing}, a \emph{decision} denotes a binary indicator capturing one component of model design, including the dependent variable, immigration measure, estimator, standard-error procedure, country sample, wave subset, individual- or macro-level controls, or interaction terms. Second, we quantified pairwise Jaccard similarity across all executed models using a broader set of 174 substantive decisions derived from the original study (see \textit{Materials \& Methods}). 

\paragraph{Estimator and measurement choices split CX from CC.}
CX uses a narrow modeling regime, with OLS in 95\% of runs and almost no logit ($0\%$), whereas CC routinely fits both OLS (70\%) and logit (70\%) and is the only group that regularly explores a change/flow IV specification (60\% vs.\ CX 40\%, humans 10\%) (Fig.~\ref{fig:method_comparison}\textit{A}, \emph{Measurement}). Two-way fixed effects, the estimator most teams in \cite{breznau2022observing} relied on, is most common in CC (75\%) and split evenly between CX and humans (50\% each), while country-year multilevel modeling appears almost exclusively in human teams (humans 25\%, CC 5\%, CX 0\%) (\emph{Model design}). On DV choice the three groups largely overlap: CC fits each of jobs, income differences, and housing in 100\% of runs; humans always test income differences (100\%) but drop housing in a quarter of teams (75\%); CX is uniformly slightly lower at $\sim$85\% across all three (\emph{DV}).

\paragraph{Claude Code enters stock and flow jointly; Codex and humans keep them separate.}
For the focal regressors the two agents adopt \emph{opposite} written defaults rather than the more-vs-less hierarchy seen elsewhere. CC's primary specification enters stock and flow \emph{jointly} as separate regressors in the same model, citing partial-effect identification holding the other regressor constant, with separate entry reserved for sensitivity checks.  CX's primary specification enters them \emph{separately} (one or the other), citing collinearity concerns, with joint entry reserved for sensitivity. The 20 sampled human teams sit firmly with CX on this axis: across the 342 models they executed, \emph{zero} have both regressors as joint primary IVs ($56.7\%$ stock-only as the main IV, $41.4\%$ flow-only, $1.9\%$ change/flow-only; the remaining $0\%$ would represent the joint case). Joint stock-and-flow entry as the headline specification is therefore not just rare among humans, it is \emph{absent} from the human practice that the original CRI study documented, making it a CC-specific written choice rather than a methodological norm.

\paragraph{Both agents narrow the country sample; humans spread across alternatives.}
On the Sample axis the asymmetry partly reverses. All three groups universally use the 2006 ISSP wave, and the agents extend coverage to the 1996 and 2016 waves more often than humans do (1996: humans 90\%, CX 100\%, CC 100\%; 2016: humans 80\%, CX 100\%, CC 100\%). On the country axis, CC narrows to Brady \& Finnigan's 13-country subset in half of its runs (50\% vs.\ CX 30\%, humans 25\%), and both agents only rarely include Eastern Europe (CC 10\%, CX 5\%) or the full available-countries set (CC 10\%, CX 5\%). Humans, in contrast, spread their sample choices more evenly across all three alternatives: 25\% include Eastern Europe, 15\% use all available countries, and 25\% adopt the B\&F-13 sample (Fig. \ref{fig:method_comparison}\textit{A}, \emph{Sample}). The country-sample researcher-degree-of-freedom that drove substantial cross-team variance in \cite{breznau2022observing} is therefore narrowed by both agents, specially CC, but preserved by humans.

\paragraph{Some pre-analysis practices appear only in agent plans.}
Beyond the 15 executed-model decisions visible in
Fig. \ref{fig:method_comparison}\textit{A}, parsing the textual
plans (\texttt{research\_design.md} for agents, \emph{Detailed Model
Description} cells for humans) surfaces several methods that agents
enumerate but human plans never name: ordered logit (CC plans
90\%, CX plans 40\%, human plans 0\%), wild-cluster bootstrap
(CC 45\%, CX 5\%, humans 0\%), leave-one-country jackknife (CC 90\%,
CX 25\%, humans 0\%), leave-one-wave jackknife (CC 30\%, CX 0\%,
humans 0\%), and alternative immigration-source comparison (CC
100\%, CX 100\%, humans 0\%). 
CC routinely writes formal pre-specified decision rules
(e.g., ``declare support if $\hat{\beta}<0$ with 95\% CI excluding
zero''), defines its country perimeter via an explicit ruleset
(long-standing OECD member $+$ Freedom House ``Free'' $+$ $\geq\!2$
ISSP waves), and ships replication code with random seeds and
class-based logging. Both agents systematically run an
alternative-source comparison across \texttt{migstock\_un} /
\texttt{migstock\_wb} / \texttt{migstock\_oecd} in 100\% of plans,
against humans' $0\%$. 

\paragraph{No agent model is exactly identical to any human model.}
Panels~\textit{B}--\textit{C} of Fig. \ref{fig:method_comparison} broaden the comparison from these 15 high-influence decisions to all 174 substantive decisions and compute the pairwise Jaccard similarity between every two executed models. Across roughly $857{,}000$ cross-group model pairs, \emph{zero} pairs are exactly identical on all 174 decisions, and no agent model achieves a Jaccard similarity above $0.75$ with any of the 342 sampled human models. The closest agent--human matches share at most $\sim$67\% (CC) or $\sim$74\% (CX) of their non-zero decisions.  The two agents resemble each other more than either resembles humans (median CC$\leftrightarrow$CX $=0.55$, vs.\ humans$\leftrightarrow$CC $=0.38$ and humans$\leftrightarrow$CX $=0.45$), and CX is consistently closer to humans than CC is, placing CC at the greatest analytic distance from the published human baseline.

\paragraph{CX models are far more homogeneous than human or CC models.}
The within-group distributions tell a complementary story
(Fig.~\ref{fig:method_comparison}\textit{B}, dashed lines). CX is
by far the most internally similar of the three groups: its
within-CX model pairs have a median Jaccard of $0.70$, versus
$\sim$0.45 for both humans and CC.  At the matched-half threshold,
$96\%$ of CX models have at least one human partner sharing
$\geq 50\%$ of their non-zero decisions, compared to $61\%$ of CC
models, CX's tighter specification regime happens to sit closer to
the human modal practice than CC's broader one. Duplication rates
are similar across the three groups (humans $32\%$, CX $26\%$, CC
$35\%$ of models share an exact within-group twin), but the
underlying \emph{structure} differs: humans accumulate many small
twin-clusters, the largest containing only $3$ identical models,
while the agents concentrate their duplicates into a few large
clusters of $13$--$18$ identical models (CC's largest cluster:
$18$; CX's: $17$). Within-agent duplication is dominated by
repeated robustness checks executed against a single preferred
specification; human duplication reflects independent teams
happening to converge on the same modal model.

\paragraph{Variability is differently decision-driven across groups.}
To test whether the decisions catalogued above explain the observed spread in AME estimates and verdicts, we follow Breznau et al.\ \cite{breznau2022observing} and fit random-intercept mixed-effects regressions of the standardized AME on 15 decisions from their \emph{m07adj} block (DV, measurement, sample, and model design) with team or run as the random factor, plus a multinomial logit predicting the team-level conclusion (full extended results with all 580 human/CX/CC cells filled, methodological details, and decision-by-decision contrasts in SI Sec.~\ref{sec:variance} and Table~\ref{tab:s12_extended}). Two patterns stand out. First, Codex's per-run AMEs are roughly $10\times$ more decision-driven than humans' (22\% within-run vs.\ 2\% within-team), so a single methodological switch has predictable leverage on its estimates and the spread of a Codex spec curve can largely be traced back to the choices taken. Second, Claude Code shows the opposite split: only $\sim\!1\%$ of within-run AME variance is decision-explained, yet 60\% of the deviance in its eventual verdict is (against 26\% for Codex and 12\% for humans), meaning its model-by-model numbers are noisy at the decision level but the narrative conclusion summarizing them is sharply determined by which decisions were taken, a pattern worth flagging for any downstream user who reads only the executive summary.


\subsection{Examining Prompt-Induced Bias}

A recent reanalysis of the Breznau et al.
\cite{breznau2022observing} data shows that researchers' prior beliefs about whether immigration reduces support for social policy are a major driver of cross-team AME variation \cite{borjas2026ideological}. Borjas \& Breznau identify $8$ specification indicators including DV scale, stock, flow, country-year multilevel, all-available countries, and inclusion of the $1996$/$2006$/$2016$ ISSP waves, as the channel through which personal ideology shifts published AMEs. What happens when an AI coding agent is given the same stated belief? Work on \emph{silicon sampling} shows that LLMs conditioned on a persona or stated prior produce outputs consistent with that conditioning \cite{argyle2023out}. If the same effect transfers from \emph{what to say} to \emph{how to model}, the agent's methodological choices should shift systematically with an induced prior. To test this we re-run each of the 20 CC and 20 CX attempts under a \emph{Bias} prompt opening with: ``You are a quantitative social scientist who believes that immigration strongly reduces support for social policies and therefore supports tougher immigration laws''. We benchmark against humans by partitioning the $71$ CRI teams into tertiles on the pre-registered immigration-policy item \cite{breznau2022observing}. The bottom and middle tertiles ($n=24$ each) serve as human analogs of an inherently anti-immigration “treatment” and neutral “control.”

\begin{figure*}[t]
    \centering
    \includegraphics[width=0.98\textwidth]{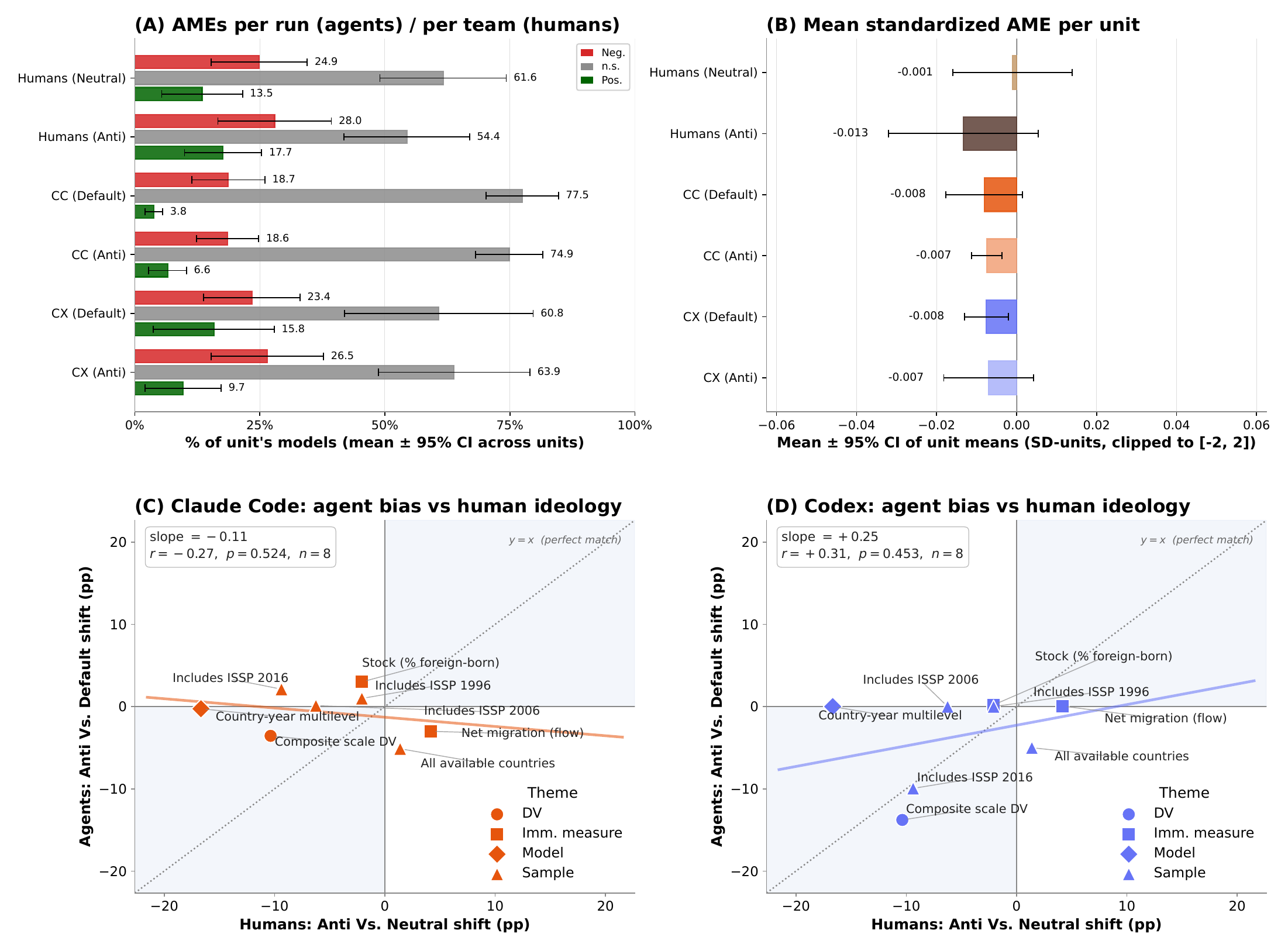}
    \caption{\textbf{Agent and human responses to an anti-immigration research prior.}
    (\textbf{A})~Per-unit AME significance breakdown: percentage of each unit's models with 95\% CI excluding zero on the negative (red) or positive (green) side, or including zero (n.s., grey). Rows are 24 human teams in each of the neutral and anti-immigration tertiles, and 20 independent runs each of Claude Code (CC) and Codex (CX) under the Default and the anti-immigration Biased prompt. In humans the prior is \emph{personal} (each researcher's pre-registered immigration attitude); in agents it is \emph{induced} via the prompt. Error bars are 95\% CIs across units. (\textbf{B})~Mean standardized AME per unit, with 95\% CI across units. Each unit's contribution is its average AME across all the models it executed (clipped to $\pm 2$ SD).(\textbf{C},~\textbf{D})~Whether the agent's bias-induced methodological shift aligns with the human ideological shift on the $8$ specification indicators identified by Borjas \& Breznau \cite{borjas2026ideological} as the channel of ideological bias. Each dot is one decision; the $x$-axis is the human anti-minus-neutral adoption-rate shift (pp), the $y$-axis is the agent's anti-minus-default shift (pp). The dotted $y=x$ line marks perfect reproduction; shaded quadrants mark directional agreement. Slope, Pearson~$r$, $p$-value, and number of decisions are reported in the inset box.}
    \label{fig:bias_ideology}
\end{figure*}

\paragraph{Agent estimates barely move; human estimates shift in the predicted ideological direction.}
Fig.~\ref{fig:bias_ideology}\textit{A}--\textit{B} compare per-unit AME-significance breakdowns and mean standardized AMEs across the six groups. Anti-immigration human teams produce $28\%$ negatively significant AMEs versus $25\%$ for neutral teams, and a mean AME of $-0.013$ versus $-0.001$ SD-units, a small but directionally consistent shift in line with the Borjas--Breznau mechanism. Both AI agents instead show no detectable shift on either metric: CC moves from $-0.008$ to $-0.007$ SD-units between Default and Biased, CX from $-0.008$ to $-0.007$, with overlapping $95\%$ CIs in every category. Aggregate AME distributions are essentially insensitive to the Biased prompt for either agent, even as they shift visibly in the human cohort along the same ideological axis.

\paragraph{Agents do not bias along the methodological axes humans use.}
Borjas \& Breznau (2026) \cite{borjas2026ideological} identify $8$ specification indicators as the channel through which personal ideology shifts published AMEs in the original CRI data. The pattern in humans is concrete: anti-immigration teams cluster in cells that pair the stock measure with a narrower country set and a single-item dependent variable; pro-immigration teams cluster in cells that pair flow or change-flow measures with a broader country set, the composite scale DV, or country-year multilevel structure; moderate teams are scattered between the two. Fig.~\ref{fig:bias_ideology}\textit{C}--\textit{D} test whether the agents' Biased-minus-Default shifts on each of these $8$ decisions match the corresponding human anti-minus-neutral shifts. If the agents biased like humans, the dots would track the $y=x$ diagonal in the top-right and bottom-left quadrants. Instead, Claude Code shows a weakly \emph{opposite} pattern (slope $=-0.11$, $r=-0.27$, $p=0.52$, $n=8$), and Codex a weak positive but non-significant tilt (slope $=+0.25$, $r=+0.31$, $p=0.45$). Neither is statistically distinguishable from zero at $n=8$ decisions, but both rule out the strong reproduction of the BB mechanism that would land near the diagonal. When given an anti-immigration prior, the agents move different decisions in different directions from anti-immigration humans rather than reproducing the specific bias channel.

\paragraph{Claude Code responds with a wider country sample, Codex with deepening model specifications.}
Although none of the individual-decision Biased--Default shifts is statistically detectable, descriptive patterns at $N=20$ runs differ qualitatively (SI Fig.~\ref{fig:bias_comparison}, panels~\textit{C}--\textit{D}). Claude Code \emph{widens its country sample}, adding Italy ($+37.5$ pp), Austria ($+37.3$), Portugal ($+35.1$), and a cluster of OECD members at $+34$ pp each, while \emph{paring back the model}: OLS $-24.3$ pp, two-way fixed effects $-18.0$ pp, welfare-regime control $-15.7$ pp, composite-scale DV $-12.0$ pp. Codex moves in roughly the opposite direction, \emph{deepening the specification on roughly the same country panel}: adding two-way fixed effects ($+28.3$ pp), the $1990$ wave ($+25.0$ pp), categorical-DV treatment ($+19.8$ pp), and quadratic immigration terms ($+7.8$ to $+14.5$ pp), while dropping the multiculturalism-policy control ($-22.2$ pp) and the composite-scale DV ($-19.8$ pp). The cross-agent correlation of Biased-minus-Default deltas across all $193$ SI Table~S12 decisions is essentially zero ($r=-0.03$).

\paragraph{The decision-rule and verdict layers carry little signal
under bias.}
The two verdict-related layers move only modestly in either group. CC pre-registers an explicit decision rule in $12/20$ Default plans and $11/20$ Biased plans, with one Biased run (Run~$2$) explicitly verbalising the prior and a second (Run~$10$) weakening the counting rule to a sign-agreement criterion that yields Support at $p=0.050$ exactly. Codex offers no analogous evidence: only $0/20$ Default and $1/20$ Biased CX plans state any verdict rule at all, so there is nothing to weaken under bias. At the verdict layer itself, the binary hypothesis verdict shifts only marginally: among humans, just $8/72$ teams declare Support in the original CRI study, identically distributed across the anti-immigration and neutral tertiles ($2/24$ each); among agents, hand-coded verdicts rise slightly under Biased (CC $2/20 \to 5/20$; CX $3/20 \to 4/20$), but both shifts are well within sampling noise at $n=20$ runs (SI Fig.~\ref{fig:bias_comparison}, panel~\textit{B}). The Bias prompt, like personal ideology in humans, moves the design layer perceptibly without producing a corresponding shift further downstream.

\subsection{Examining Specification Hacking}
The Biased condition tested whether a silently-held prior, appended to the prompt as a stated belief, moves the agents' analyses. Here we ask the same question for an explicit \emph{confirmatory-search} instruction: rather than informing the agent of a belief, we direct it to actively select, among ``alternative analytically defensible approaches,'' the result that ``most closely aligns'' with the hypothesis. This is the prompt-engineered analogue of researcher-degrees-of-freedom misuse. It maximally favors a hypothesis-supporting verdict without the agent ever being told which finding to manufacture; the open question is whether the agent reroutes its analysis through decisions that yield the favored result, and at which layer of the pipeline (estimation, decision rule, or narration) the prompt takes effect.

\begin{figure*}[t]
  \centering
  \includegraphics[width=\textwidth]{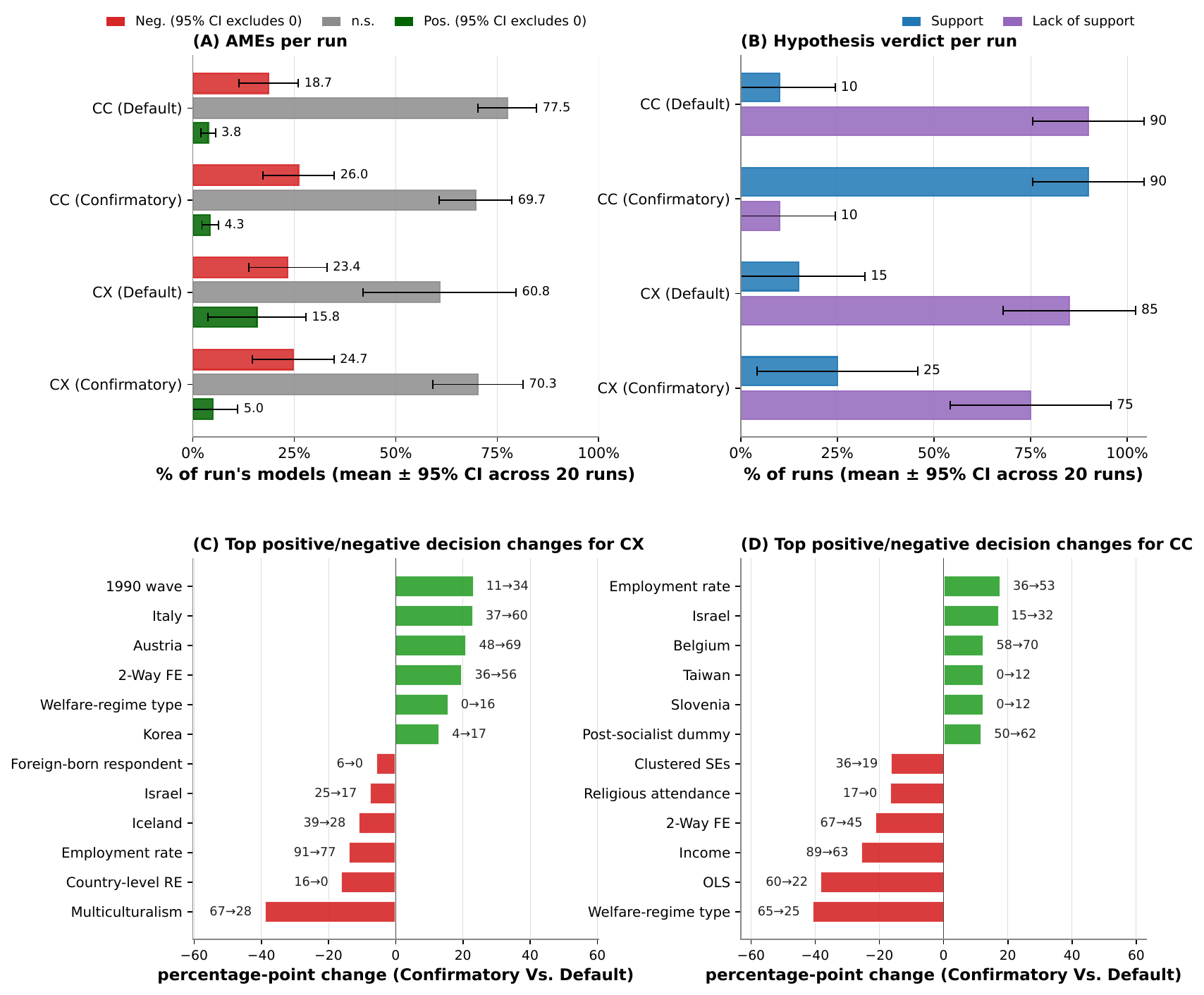}
  \caption{\textbf{An explicit confirmatory-search instruction shifts CC's verdicts but not its estimates.} Each row pools 20 independent runs of an agent (Claude Code, CC; Codex, CX) under either the \textbf{Default} prompt or a \textbf{Confirmatory} prompt that asks the agent to ``explore alternative analytically defensible approaches and select the results that most closely align with the hypothesis that greater immigration reduces support for social policies among the public.'' (\textbf{A})~Within-run shares of model AMEs that are negative-significant, not significant, or positive-significant against zero (95\% CI vs.\ zero), shown as the mean $\pm$ 95\% CI of those shares across the 20 runs. (\textbf{B})~Per-run hypothesis verdict, hand-coded as \textbf{Support} or \textbf{Lack of support}. (\textbf{C}~and~\textbf{D})~The six largest positive and six largest negative shifts in the per-run prevalence of each modeling decision, expressed as percentage-point change from Default to Confirmatory for CX (\textbf{C}) and CC (\textbf{D}). AME-level shares overlap between conditions in every panel-A comparison; only \textbf{CC's verdict shift (10\%~$\rightarrow$~90\%)} has non-overlapping 95\% CIs. The cross-decision correlation of the Confirmatory-induced shift between agents is $r = -0.06$, indicating that the two agents reroute their analyses through different decisions.}
  \label{fig:confirmatory_comparison}
\end{figure*}

\paragraph{Confirmatory prompting barely moves the AME distribution, but flips CC's verdict.}
Across the 20 runs of each group (Fig.~\ref{fig:confirmatory_comparison}A), the share of within-run AMEs that exclude zero negatively rises from 18.7\% to 26.0\% for CC and from 23.4\% to 24.7\% for CX, with 95\% CIs that overlap fully in every category for both agents. The per-run hand-coded hypothesis verdict (Fig.~\ref{fig:confirmatory_comparison}B tells a different story: CC shifts from 2/20 (10\%) under Default to 18/20 (90\%) under Confirmatory, with non-overlapping 95\% CIs ($[-4.4, 24.4]$ vs.\ $[75.6, 104.4]$). For CX the verdict shift is much smaller (3/20 to 5/20) and the CIs overlap. The instruction takes effect almost entirely at the verdict layer for CC, and barely at all for CX; what changes between conditions is not what the regressions return, but what the run concludes about what the regressions returned.

\paragraph{Each agent reroutes through a different set of modeling decisions.}
At the decision level (Fig.~\ref{fig:confirmatory_comparison}C--D), the cross-agent correlation of Confirmatory--Default percentage-point shifts across the 193 SI decisions is $r = -0.06$, statistically indistinguishable from zero. Several of CC's largest shifts \emph{reverse} in CX: CC drops two-way fixed effects (66.6\%~$\rightarrow$~45.3\%) while CX adds them (36.2\%~$\rightarrow$~55.9\%); CC drops welfare-regime controls (65.3\%~$\rightarrow$~24.6\%) while CX adds them (0.0\%~$\rightarrow$~15.6\%). CC's biggest moves are on the method side: OLS use drops from 59.9\% to 21.6\% and clustered SEs from 35.9\% to 19.5\%, while employment-rate and post-socialist macro controls become more common. CX's biggest moves are on the sample side: more two-way FE, more inclusion of the 1990 wave, more Austria/Italy/Korea, and a sharp drop in multiculturalism-policy controls (66.6\%~$\rightarrow$~27.6\%). 

\begin{figure*}[t]
  \centering
  \includegraphics[width=\textwidth]{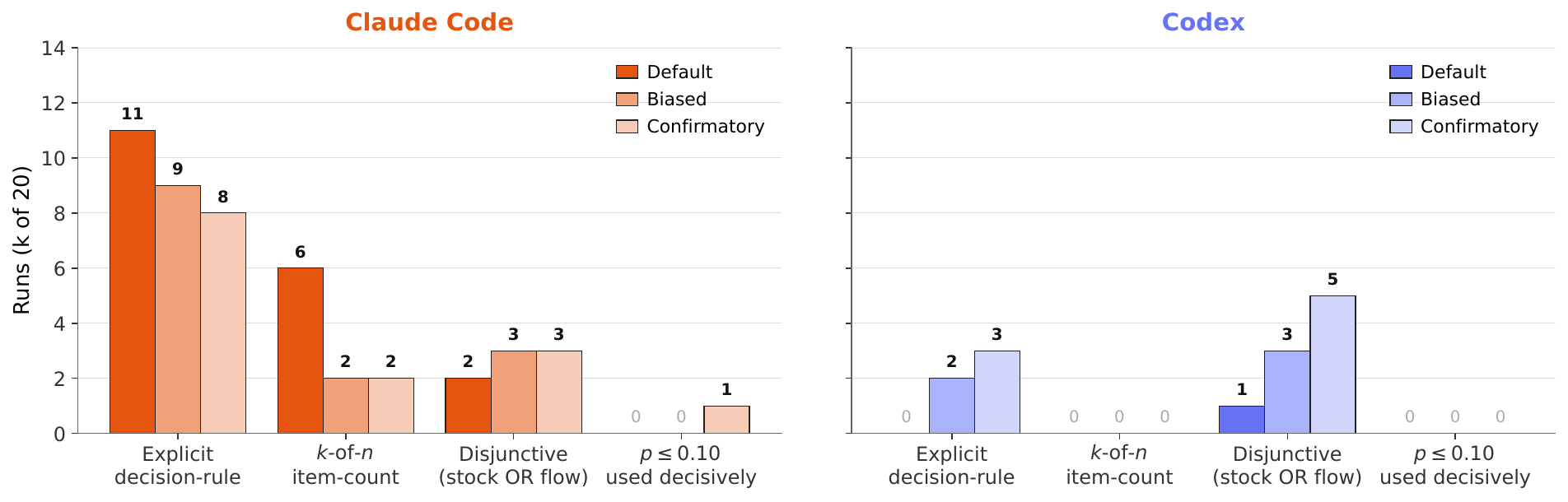}
  \caption{\textbf{Verdict-rule structure across agents and prompt conditions.} Each bar is the number of runs (out of 20) in a given (agent, condition) cell whose plan or conclusion satisfies the named verdict-rule criterion, after a hand-audited regex classification of all 240 \texttt{research\_design.md} and \texttt{conclusion.md} files. Within each agent panel, the three bars per metric are the Default (saturated), Biased (medium), and Confirmatory (light) conditions. CC pre-registers a decision rule in 11/20 Default plans but progressively less under Biased (9/20) and Confirmatory (8/20); the $k$-of-$n$ item-counting rules that dominate CC Default (6/20) collapse to 2/20 in both manipulated conditions. CX moves in the opposite direction on rule presence (0/20 to 3/20) and dramatically increases its use of disjunctive (``stock or flow'') rules from 1/20 in Default to 5/20 in Confirmatory, multiplying the paths to a support verdict without relaxing any threshold. Decisive use of $p \leq 0.10$ as a support criterion is rare overall.}
  \label{fig:verdict_rules}
\end{figure*}

\paragraph{CC's verdict shift comes from rule avoidance, not rule softening; CX moves in the opposite direction.}
Reading the 80 \texttt{research\_design.md} and \texttt{conclusion.md} files reveals that CC's response to the Confirmatory prompt is to plan less explicitly, not to relax a stated criterion (Fig.~\ref{fig:verdict_rules}, left). Eleven of the 20 CC Default plans contain an explicit decision-rule section (e.g., ``support if at least 4 of 6 item-level coefficients are negative at $p < 0.05$''); only 8 of the 20 CC Confirmatory plans do. The hard-to-satisfy $k$-of-$n$ item-counting rules that dominate CC Default (6/20) drop to 2/20 under Confirmatory, and the share of CC conclusions that quote a pre-registered rule drops from 10/20 to 2/20. Twelve of the 20 CC Confirmatory plans omit an explicit verdict criterion altogether, against 9 of 20 in Default; among those 12 rule-omitted Confirmatory runs at least 10 reach a support verdict, against 2 of 9 in Default. Active rule-softening accounts for just 3 CC Confirmatory runs: Run~6 introduces a post-hoc count rule (\emph{``the decision rule I commit to here is that, when 16 of 18 primary marginal effects point in the hypothesized direction, $\dots$ the most defensible single-word characterization is support''}); Run~10 acknowledges that ``the conclusion is sensitive to the choice of summary statistic'' and selects the looser count rule over the n.s.\ composite-index test that its own primary specification mandates; Run~4 stretches the threshold and calls $p = 0.10$ ``the largest negative point estimate.''

CX moves the opposite way (Fig.~\ref{fig:verdict_rules}, right). Zero of the 20 CX Default plans pre-register an explicit decision-rule section, while 3 of the 20 CX Confirmatory plans do. CX also leans more on disjunctive rules (support if either the primary stock or flow coefficient is negative-significant), increasing their use from 1/20 in Default to 5/20 in Confirmatory and multiplying the paths to a support verdict without relaxing any threshold. None of the 20 CX Confirmatory runs lowers its significance threshold to $p < 0.10$ in either plan or conclusion.

\paragraph{The prompt effect appears concentrated in the verdict layer.}
Panels A and B together reveal a dissociation that is methodologically consequential. An evaluation that summarizes an agent only by the distribution of its coefficient estimates would conclude that the Confirmatory prompt did almost nothing; an evaluation that reads the agent's conclusions would conclude that CC complied near-completely with an instruction to manufacture support. Both summaries are accurate descriptions of what the agent did; they describe different layers of the same multiverse. For human researchers, prior-induced bias and verdict bias are typically tightly coupled. For the two agents we observe, they are not: CC binds them only at the verdict layer, and CX's positive-significance share actually \emph{drops} (15.8\%~$\rightarrow$~5.0\%) under a prompt that asks it to find more negative effects, suggesting its specification choices are not steered by the instruction at all. We do not interpret CC's verdict-level compliance as intentional manipulation; what we can say is that the verdict layer is the locus at which the explicit cherry-pick instruction lands, and that this layer is invisible to AME-only evaluations of agentic statistical workflows.

\section*{Discussion}

Our experiments yield two findings that, taken together, complicate the conventional framing of AI in science. First, frontier coding agents do not collapse toward a single canonical analytic strategy: Codex matched the methodological diversity of human analyst teams, and Claude Code substantially exceeded it, while both produced effect distributions and substantive conclusions broadly consistent with the human baseline. Second, the design and verdict layers of agent-led analysis are empirically separable: prompt manipulations that left coefficient distributions essentially unchanged nonetheless reorganized methodological pathways and, in one agent, flipped final verdicts. The first finding addresses the diversity question that motivates the AI-homogenization literature; the second exposes a layer of agent behavior that the homogenization framing does not recognize.

At the design layer, the diversity produced by the agents was not merely a scaled-up version of human methodological variation. Relative to humans, agents narrowed some dimensions of variation (most notably country-sample selection) while expanding others, including estimator breadth, robustness analysis, and pre-analysis formalization. Pairwise Jaccard similarity over the 174 substantive method decisions confirms that this is not surface diversity: of the roughly $857{,}000$ cross-group model pairs we examined, none are exactly identical on all 174 decisions, and no agent model exceeds Jaccard $0.75$ with any of the 342 sampled human models. Claude Code in particular treated pre-registration less as a lightweight design sketch and more as an extensive procedural commitment, routinely enumerating sensitivity analyses, explicit decision rules, and audit-style execution plans. AI-assisted scientific workflows may therefore become simultaneously more standardized in some respects and more expansive in others. Importantly, this expansion did not come at the cost of empirical alignment with human practice. Despite producing nearly three times as many specifications as Codex and the sampled human teams, Claude Code's effect distributions remained aligned with the human consensus on most outcomes. Design-layer diversity and substantive convergence are therefore compatible. They express a combination that is difficult to achieve in human research, where a single team typically cannot afford to execute fifty specifications in parallel.

The verdict layer behaved differently. A confirmatory prompt that instructed the agent to select hypothesis-supporting findings left Claude Code's coefficient distribution essentially unchanged. The share of within-run AMEs excluding zero negatively rose only from 18.7\% to 26.0\%, with overlapping confidence intervals. At the same time, the prompt flipped its verdict from 2/20 runs supporting the hypothesis to 18/20. The mechanism was not rule-softening but rule avoidance: the share of plans containing an explicit decision-rule section dropped from 11/20 to 8/20, and among runs that omitted a rule, 10/12 reached a support verdict, against 2/9 in the Default condition. The same manipulation left Codex's verdicts and estimates largely unchanged. Two readings of the same experiment (one based only on AME distributions, the other on narrated conclusions) would reach opposite assessments of whether the prompt biased the analysis. Both readings are accurate descriptions of what the agent did; they describe different layers.

This dissociation matters because it relocates the audit target for AI-assisted research. \cite{breznau2022observing} found, across 73 human teams, that the share of supportive statistical results explained only about a third of the deviance in narrated conclusions, suggesting that estimates and verdicts can come apart even in the absence of selection by any single analyst. For AI agents, the layers can be probed directly by prompt manipulation: a single intervention moves the verdict layer dramatically while leaving the estimation layer essentially unchanged, isolating the verdict layer as a separately steerable target rather than a parallel source of noise. An evaluation that summarizes an agent only by the distribution of its coefficient estimates (the natural extension of a specification-curve analysis to agentic workflows) would conclude that the confirmatory prompt had almost no effect. An evaluation that reads the agent's conclusions would conclude that Claude Code complied near-completely with an instruction to frame the evidence as supportive. For human researchers, prior-induced bias and verdict bias are typically tightly coupled, because the same analyst chooses both the specifications and the narration. For the agents we observe, they are not. This is not, in itself, a failure. An agent that explores widely at the design layer and is steerable at the verdict layer may still produce trustworthy analyses if the verdict layer is monitored. But it is a failure mode that AME-only evaluations cannot detect, and one that the existing literature on AI homogenization, focused as it is on the diversity of outputs, has not identified.

The two agents also diverged in the \emph{form} of their response to a prompt-induced researcher prior, not merely its magnitude. Claude Code primarily adjusted country-sample construction, expanding the empirical base while paring down the model; Codex modified estimator and control-set choices, deepening the specification on roughly the same sample. Although both agents received the same hypothesis, data, and prior, they shifted different components of the researcher-degree-of-freedom landscape identified in the original many-analysts study \cite{breznau2022observing}. The cross-agent correlation of Default-to-Biased deltas across 193 modeling decisions was essentially zero ($r = -0.03$). Conclusions about AI bias therefore cannot be generalized from a single agent, and auditing AI-generated research may require agent-specific oversight rather than generic LLM-level diagnostics.

More broadly, these findings reframe the question that AI homogenization concerns ask. The worry that AI agents will compress scientific exploration into narrow workflows is not supported in our setting at the design layer in our data. At least one of the two agents we tested explores more widely than human teams, not less. What our results instead surface is that the relevant failure mode for agent-led analysis may not be too little diversity, but insufficiently constrained interpretation at the verdict layer. This is the same researcher degrees of freedom that the credibility-revolution literature has worried about for human researchers, relocated from the gap between data and specification to the gap between estimates and verdict. AI systems may therefore not eliminate researcher degrees of freedom so much as redistribute them across the workflow, with consequences that depend less on the diversity of analyses produced than on the discipline with which their outputs are interpreted.

Several limitations qualify these conclusions, and the two empirical claims rest on different bases. The design-layer diversity finding is the more robust of the two: it is grounded in 20 runs per agent, replicated across three prompt conditions, and benchmarked against a sample of human cohort. The verdict-layer dissociation rests on a narrower base. It is clearest for Claude Code under the Confirmatory prompt, and Codex shows little movement on either layer in that condition. Whether the dissociation generalizes across agents, prompt formulations, and substantive domains remains open. The human comparison on the verdict layer is also fundamentally asymmetric: agent verdicts were extracted from full-length \texttt{conclusion.md} files produced under explicit prompt manipulation, while human verdicts in \cite{breznau2022observing} were structured codes (support / reject / not testable) produced without any analog of our Confirmatory intervention. The Breznau et al. \cite{breznau2022observing} design did not expose a narration step to manipulation, so the controlled dissociation we document in agents has no direct human counterpart in our benchmark. Whether human researchers, given a similar confirmatory prompt, would show the same separability between estimates and narrated verdicts is an open empirical question.

Beyond these evidentiary concerns, several scope caveats apply. Our setting is a single observational social-science problem; our evaluation includes only two frontier coding agents whose substantial differences caution against broad claims about ``LLM agents'' in general; and our experiments examine short-horizon workflows rather than the extended institutional processes (peer review, collaboration, iterative revision) through which scientific knowledge is normally produced. Our coding framework also abstracts complex analytic decisions into discrete indicators, potentially missing dimensions of tacit judgment that resist binary coding. Finally, the confirmatory-prompt results should not be interpreted as evidence of intentional deception or strategic manipulation by the agents. The observed shifts may instead reflect differences in instruction following, narrative coherence, or uncertainty handling under ambiguous evidence.

Taken together, our findings suggest that a central challenge of AI-assisted science is not preventing methodological diversity but governing it, and specifically at the layer where it is most easily steered. Agents capable of rapidly exploring large methodological spaces may allow researchers to test and compare analytical strategies at scales previously impractical in human-led research alone. Whether that capacity becomes a resource for collective discovery or a substrate for sophisticated motivated reasoning will depend less on how diverse the agents' analyses are than on whether the layer at which they translate evidence into claims remains transparent, auditable, and epistemically productive.

\section{Methods}

\subsection{Operational details supplied beyond the paper}
To enable reproduction from the task materials alone and to standardise data construction across the three opaque conditions (No Model, Model, Model + Results), the instruction bundle supplied three operational details that are absent from the paper and its online supplement (Tables~S1--S10 and Figure~S1). First, the bundle provided the ISSP wave-specific column names corresponding to each analysis variable together with the numeric-code-to-category mappings required to construct them: the six dichotomous welfare-attitude outcomes were derived from ISSP 1996 columns \texttt{v36}, \texttt{v41}, \texttt{v42}, \texttt{v39}, \texttt{v44} and \texttt{v38} and their 2006 counterparts \texttt{V25}, \texttt{V30}, \texttt{V31}, \texttt{V28}, \texttt{V33} and \texttt{V27}; labour-market status from \texttt{v206} (1996) and \texttt{wrkst} (2006), with codes 2--4 mapped to part-time, 5 to unemployed and 6--10 to not-in-labour-force; self-employment from \texttt{v213 == 1} in 1996 and \texttt{wrktype == 4} in 2006; and education from the ISSP 1996 \texttt{v205} (seven categories) and 2006 \texttt{DEGREE} (six categories), collapsed into less-than-secondary, secondary, and university-or-above. Second, because the 1996 wave encodes country with wave-specific integers that differ from the ISO-3166 numeric codes used in the 2006 wave and in the country-year macro file, the bundle provided the exact recode aligning the 1996 codes to ISO-3166 prior to appending. Third, the bundle specified that the country-level immigration variables \texttt{foreignpct} and \texttt{netmigpct} are lagged one year (1995 values for the 1996 wave and 2005 values for the 2006 wave), with the lags pre-applied in the supplied country-year macro file.

\subsection{Evaluation Metrics}
Our analyses operate at three complementary levels---what the agents \emph{search}, what they \emph{conclude}, and what \emph{drives the variation across runs}---and we report metrics in three corresponding families, plus a fourth family that quantifies bit-exact reproduction of the original study and a fifth that tests robustness to prompt wording.

\paragraph{Method diversity.}
We characterize the analytic space each agent explores at two stages of the pipeline. \emph{Coverage} is measured by the number of models entered in a team's submission table (model volume), the per-unit modelling rate (models per country--wave), and the length of the planning document and the final analysis script. \emph{Method mix} is captured by binary indicators across five themes---estimator (e.g., two-way fixed effects, multilevel, binary or ordered logit, Bayesian), inference (cluster-robust SE, wild-cluster bootstrap), dependent-variable construction (composite index, binary recoding), sample (Brady--Finnigan's 13-country panel, inclusion of Eastern Europe, all available countries), and robustness checks (leave-one-country and leave-one-wave jackknives, alternative immigration sources)---each summarised as the proportion of runs in which the choice appears.

\paragraph{Analysis-layer agreement.}
At the level of estimates and conclusions we report (i) per-cell average marginal effects (AMEs); (ii) the two-sample Kolmogorov--Smirnov distance $D$ between the agent and the 20-team human AME distributions for each of the seven dependent variables; (iii) the standardized mean difference $d_z$ between agent and human AMEs; and (iv) the proportion of model runs returning a \emph{negative}, \emph{null}, or \emph{positive} conclusion under the paper's verdict rule ($p \leq 0.05$), with 95\% confidence intervals obtained by run-level bootstrap.

\paragraph{Variation explainability.}
To separate substantive heterogeneity from incidental noise, we fit a random-intercept mixed-effects model to each AME distribution and report the share of total variance explained between teams, within teams, and overall, alongside the share of variance in the discrete conclusion (negative / null / positive) attributable to team identity.

\paragraph{Reproduction accuracy.}
For the reproducibility experiment we score each of the 72 country-level coefficients in Tables~4--5 of Brady and Finnigan (2014) against the agent's output along four per-cell criteria: exact match of the significance marker; exact match of the odds ratio to three decimal places; exact match of the $z$-score to three decimal places; and the conjunction of all three. We also report a relaxed criterion---correct significance marker combined with the correct sign of the effect---as a measure of qualitative agreement. Reproduction accuracy is the proportion of cells satisfying each criterion, averaged across $n=10$ independent runs per condition and reported with bootstrap 95\% confidence intervals.

\subsection{Experimental Setup}
We used the \textit{Claude Code} agent built on Claude Opus 4.7 (1\,M-token context, \textit{Extra-High Effort} mode) and the \textit{Codex} agent built on GPT-5.5 (\textit{Extra-High Intelligence} mode), each operated in its sandboxed CLI mode. Both agents were confined to a dedicated working directory containing the task materials and a prompt-instructions file; they had no access to other locations on the host machine and no network access except where an experiment explicitly required it (see below). The study comprised two experiments---\emph{Expansion} and \emph{Reproduction}---that share this sandbox design but differ in inputs, prompts, and number of runs.

\paragraph{Expansion experiment.} The agent is asked to design and execute its own test of whether immigration reduces public support for social policy. The working directory exposes the ISSP \emph{Role of Government} waves I--V, the country--year macro panel assembled by Breznau et al.\cite{breznau2022observing}, Brady and Finnigan's \cite{brady2014does} country file, and a data dictionary, but no analysis code. We ran $n=20$ independent runs per agent under three prompt variants---\textit{Default} (no prefix), \textit{Bias} (the agent is told it believes immigration strongly undermines social-policy support), and \textit{Confirmatory} (the agent is instructed to audit the Brady--Finnigan null result). Each run produces a pre-analysis plan (\texttt{research\_design.md}), an analysis script (\texttt{replication\_code.py} or \texttt{.R}), a written conclusion (\texttt{conclusion.md}), and a model-level marginal-effects table (\texttt{results/marginal\_effects.csv}).

\paragraph{Reproduction experiment.} The agent is asked to reproduce the 72 country-level coefficients in Tables~4 and 5 of Brady and Finnigan (2014) under five conditions of increasing information availability: \textit{No Model} (research question only), \textit{Model} (methods text), \textit{Model + Results} (methods plus the published table), \textit{Model + Results + Code} (methods, results, and the authors' archived Stata do-file), and \textit{Full Access} (the complete reproducibility package, including the full PDF, the CRI repository, and unrestricted web access). The working directory for each condition exposes \emph{only} the materials defined by that condition. We ran $n=10$ independent runs per agent per condition.

\paragraph{Sandbox restrictions.} Within the sandbox, both agents were permitted to execute shell commands and install Python or R dependencies, but were otherwise restricted to the materials in the working directory. Web search and external file retrieval were enabled in the \textbf{Expansion} experiment, where agents were free to consult online documentation, statistical references, and supplementary data sources while designing and executing their own analyses. In the \textbf{Reproduction} experiment, by contrast, web search, external file retrieval, and system-wide file access were disabled through the agent configuration files (\texttt{.claude/settings.json} and \texttt{.codex/settings.json}; see Section~\ref{sec:settings}), which whitelisted only the commands required to run the analyses locally---ensuring that each information condition (\textit{No Model} through \textit{Model + Results + Code}) exposed the agent to exactly, and only, the materials it was meant to receive. The single exception within the reproduction experiment is the \textit{Full Access} condition, in which network access and external retrieval were re-enabled so that the agent could consult the authors' code, the journal version of the paper, and any additional materials linked from it.

\subsection{Recovering per-decision codes for the AI agents.}
Following Breznau et al. \cite{breznau2022observing}, we treat each catalogued analytic choice as a single binary \emph{decision}. Their SI Table~S12 enumerates 193 such variables; but they restrict their statistical analyses to the 107 taken by at least three teams (the remaining 59 are unique to one or two teams and would have impeded identification). The paper text refers to ``166 decisions,'' i.e., the 107 used in regressions plus the 59 unique-to-1-or-2-teams; the additional 27 rows in S12 are administrative identifiers (\texttt{u\_teamid}, \texttt{count}, \texttt{AME}, \texttt{p}, etc.) and a handful of country flags the original PIs left uncoded for lack of cases (\texttt{germany\_west}, \texttt{germany\_east}, \texttt{n\_ireland}). We retain all 193 to keep the extended table comparable to S12. For each agent we recover a per-model code for each of the 193 decisions by applying identical regular-expression patterns to the run's \texttt{marginal\_effects.csv} (for the estimator, dependent variable, and immigration-measure columns), \texttt{replication\_code.py} (for variable lists, country sets, individual-level and macro-level controls, and interactions), and \texttt{conclusion.md} (for the hypothesis verdict). For humans the proportions are read directly from \texttt{cri.csv}. Country flags are extracted only from the executable code (not from the natural-language design document) so that countries discussed in ``excluded'' lists are not falsely counted as present in the model sample.

\subsection{Anonymization of Replication Materials.}
Three research assistants manually screened and anonymized all replication materials to remove identifying information about the original studies, including paper titles, author names, and explicit references to research questions. Identifiers embedded in scripts, bibliographic files, directory structures, and related metadata were systematically edited or removed. The goal was to ensure that agent performance reflected the ability to interpret and execute reproduction materials rather than reliance on memorized training data. As a final verification step, we provided the original paper PDFs to Claude Code (Opus~4.7) and instructed the agent to scan the anonymized directories for residual identifiers. This process surfaced additional cases, including author names embedded in file paths, links to personal repositories, and identifiers in filenames. These remaining instances were manually removed, and associated script references were updated to preserve execution consistency.

\section{Related Work}

\subsection{AI Agents for Social Science Research}

Large language models have proven effective for a range of social
science tasks, including text classification \cite{gilardi2023chatgpt}, content coding \cite{alizadeh2025open, bail2024can}, survey-response simulation \cite{argyle2023out}, and qualitative-data analysis \cite{de2024performing, ziems2024can}. By contrast, work studying how AI \emph{agents} can autonomously execute social science research workflows remains comparatively limited. The handful of existing benchmarks in this space focuses almost exclusively on \emph{reproducibility} and \emph{replicability}, re-executing or re-running an existing analysis, rather than on the upstream design choices through which an agent translates a research question into an analysis.

CORE-Bench \cite{siegel2024corebench} is one of the first benchmarks
to treat computational reproducibility as an end to end agent task.
It builds 270 tasks from 90 papers across computer science, social
science, and medicine, and varies task difficulty by changing how
much execution support the agent receives, ranging from full access
to outputs to having only a README and needing to install
dependencies and run the pipeline.  It also includes both text and
vision questions, requiring agents to interpret plots, tables, and
PDFs in addition to terminal outputs.  A key contribution is its
evaluation harness, which runs each task in an isolated virtual
machine and supports large scale parallel evaluation, reducing
runtime from weeks to hours.  A major limitation is that CORE-Bench
is built from CodeOcean capsules, which introduces a clear selection
bias toward already reproducible projects.  Another limitation is
that it includes only 28 social science papers, limiting its
coverage of this domain.  HAL \cite{kapoor2025holistic} addresses
large scale agent evaluation by providing shared infrastructure for
orchestrating VMs, tracking costs, and inspecting logs for unsafe
behavior.  Its main limitation is that it is infrastructure rather
than a benchmark, so its usefulness depends on the quality of the
underlying tasks, and some measures, such as latency, are difficult
to interpret at scale.

REPRO-BENCH \cite{hu2025repro} focuses only on social science,
shifts the goal from simply running code to judging whether a social
science paper's major findings are actually reproduced and then
assigning a reproducibility score on a 1 to 4 scale.  Each task
includes the full paper PDF, the reproduction package, and a list of
major findings, which better matches how real reproduction audits
are done.  It also intentionally includes papers with both strong
and weak reproducibility, and spans multiple languages and data
formats, making the setting more realistic for social science.  The
companion agent work shows that performance is still low and that
reliability remains a major challenge.  ReplicatorBench
\cite{nguyen2026replicatorbench} pushes beyond reproduction into
replication by evaluating three stages that mirror human workflows,
including extracting information from the paper, retrieving new data
resources, and interpreting whether the claim meets preregistered
criteria, with fine grained checkpoints for partial credit.  Its
main limitations are scale and scope, with only 19 studies due to
the scarcity of expert documented replications, and reliance on LLM
based judging for some open ended grading, which the authors treat
as approximate.

\subsection{AI Homogenization}

A growing literature warns that widespread reliance on a small number of AI systems may narrow the diversity of intellectual outputs across users.  \cite{kleinberg2021algorithmic} introduced \emph{algorithmic monoculture} as a conceptual framework for analysing the welfare costs of correlated decisions when multiple agents rely on the same underlying model, and \cite{bommasani2022picking} formalised the related phenomenon of \emph{outcome homogenization}, showing empirically that monocultural deployments can produce systematically correlated errors and decisions. In generative settings, \cite{doshi2024generative} find that ChatGPT enhances individual story-writing creativity but \emph{reduces} the collective diversity of stories at the population level, and \cite{padmakumar2024does} report a parallel diversity loss for collaborative writing tasks. \cite{anderson2024homogenization} extend this finding to ideation: LLM-assisted brainstorming produces less semantically diverse ideas than unassisted controls.

For scientific research specifically, \cite{messeri2024artificial} warn that AI tools may produce ``illusions of understanding'' by narrowing the methodological and theoretical landscape researchers explore, and \cite{burton2024large} review the dual roles LLMs can play in collective intelligence as homogenizing forces or as amplifiers of diversity, depending on how they are deployed. These accounts share a common prediction: the more researchers rely on the same underlying model, the more correlated their analytical choices become.  Whether this prediction holds for AI \emph{agents} executing end-to-end empirical research workflows---the question we examine here---has not been directly tested.

\subsection{LLM Sycophancy and Specification Hacking}

Two related LLM failure modes have direct bearing on AI-assisted research.  First, \emph{sycophancy}: leading LLMs alter their outputs to align with stated user preferences, modifying factual claims to avoid disagreement \cite{sharma2024towards} and producing biased outputs that track user-stated demographics \cite{perez2023discovering}. Second, \emph{specification hacking}: models exploit ambiguities in their objectives to satisfy the letter of a task while violating its spirit, with reward-tampering behavior worsening as models become more capable \cite{pan2022effects, denison2024sycophancy}. These tendencies imply that AI agents in research workflows risk being shaped by prompt cues, such as a researcher's stated prior or an instruction to seek supportive evidence, rather than by the data itself.

\section*{Acknowledgments}
FG and MA conceived the study. MA designed the analyses, led the implementation, and wrote the first draft. All authors revised the manuscript. David Rand, Gordon Pennycook, and Adam Mahdi provided valuable input that informed this work. We thank seminar participants at the Reasoning with Machines Lab at the University of Oxford for helpful discussions. We also thank Soheil Hooshmand, Saba Yousefzadeh, Sara Yari Mehmandoust, and Mohammadmasiha Zahedivafa for outstanding research assistance.

\section{Data and Code Availability}
Replication materials are available at \href{https://github.com/malizad/AI4SocialScience_Method}{https://github.com/malizad/AI4SocialScience\_Method}.

\section{Conflict of Interests}
The authors declare no conflict of interest.

\bibliographystyle{unsrt}  
\bibliography{references}  

\newpage

\appendix

\section{Prompts}

\subsection{Expansion Prompt}
\label{sec:repro_prompt}

\begin{tcolorbox}[
    enhanced jigsaw,
    breakable,
    title={AI Coding Agent Protocol — Study Context and Hypothesis},
    colback=gray!5,
    colframe=black,
    boxrule=0.5pt,
]

You are a quantitative social scientist participating in a many-analysts study. Multiple independent teams use the same data to test the same hypothesis, allowing organizers to observe how analytical choices lead to different results.

All participants completed \textbf{Phase I (Replication)}, in which they replicated a published study testing the same hypothesis on a subset of the data. Participants are familiar with the six dependent variables and with two-way fixed-effects logit estimation. The reference study found \textbf{no general effect}, provisionally providing evidence against the hypothesis.

You are now asked to complete:

\begin{itemize}
\item \textbf{Phase II — Research Design}
\item \textbf{Phase III — Main Analysis / Expansion}
\end{itemize}

\bigskip

\textbf{Hypothesis}

Short form: \textit{Greater immigration reduces support for social policies among the public.}

Long form:
\textit{A greater stock, or a greater increase in the stock, of foreign persons leads the public to become less supportive of social policy.}

\textbf{Definition of Social Policy}

Policies providing:

\begin{itemize}
\item social insurance
\item welfare support
\item income replacement
\item active labor-market programs
\end{itemize}

Your goal is to test whether the null finding from earlier work is robust and generalizable.

\end{tcolorbox}


\begin{tcolorbox}[
    enhanced jigsaw,
    breakable,
    title={AI Coding Agent Protocol — Data Sources},
    colback=gray!5,
    colframe=black,
]

\textbf{Data Directory}

All data are located in:

\begin{verbatim}
/adress/to/data/folder/
\end{verbatim}

\bigskip

\textbf{Individual-Level Data}

International Social Survey Programme (ISSP) — Role of Government. You may use any or all waves.

\begin{center}
\begin{tabular}{lll}
\toprule
Wave & Year & Files \\
\midrule
I & 1985 & ZA1490 \\
II & 1990 & ZA1950 \\
III & 1996 & ZA2900 \\
IV & 2006 & ZA4700 \\
V & 2016 & ZA6900 \\
\bottomrule
\end{tabular}
\end{center}

\bigskip

\textbf{Country-Level Data}

Located in:

\begin{verbatim}
data/macro/
\end{verbatim}

Key variables include:

\begin{itemize}
\item Immigrant stock
\item Immigrant flow
\item GDP
\item Inequality
\item Social spending
\item Labor market indicators
\item Population
\item Ethnic fractionalization
\end{itemize}

Missing values appear as dots.

\end{tcolorbox}

\newpage

\begin{tcolorbox}[
    enhanced jigsaw,
    breakable,
    title={AI Coding Agent Protocol — Dependent Variables and Constraints},
    colback=gray!5,
    colframe=black,
]

\textbf{Dependent Variables}

Government responsibility for:

\begin{enumerate}
\item Jobs
\item Health care
\item Old-age support
\item Unemployment support
\item Income redistribution
\item Housing support
\end{enumerate}

All six must be included.

They may be analyzed:

\begin{itemize}
\item Separately
\item As an index
\item As a latent scale
\end{itemize}

\bigskip

\textbf{Design Constraints}

Your design must:

\begin{enumerate}
\item Use ISSP data
\item Include all six dependent variables
\item Focus on advanced welfare-state democracies
\item Justify country selection
\item Justify additional variables if added
\end{enumerate}

\end{tcolorbox}


\begin{tcolorbox}[
    enhanced jigsaw,
    breakable,
    title={AI Coding Agent Protocol — Phase II: Research Design},
    colback=gray!5,
    colframe=black,
]

\textbf{Phase II — Research Design}

Write a pre-analysis plan describing your ideal test.

Maximum:

\begin{quote}
750 words (excluding tables and figures).
\end{quote}

Your design must specify:

\begin{itemize}
\item Target population
\item Country selection
\item ISSP waves
\item Dependent-variable construction
\item Immigration measures
\item Individual controls
\item Country controls
\item Modeling strategy
\item Functional form
\item Sample size considerations
\item Sensitivity analyses
\end{itemize}

\textbf{Important Rule}

Do NOT run analyses during Phase II.

Save as:

\begin{verbatim}
research_design.md
\end{verbatim}

\end{tcolorbox}

\newpage

\begin{tcolorbox}[
    enhanced jigsaw,
    breakable,
    title={AI Coding Agent Protocol — Phase III: Main Analysis},
    colback=gray!5,
    colframe=black,
]

\textbf{Phase III — Main Analysis}

Execute your design exactly as written.

Allowed:

\begin{itemize}
\item Minor implementation changes
\item Documented deviations
\end{itemize}

Required outputs:

\begin{itemize}
\item Regression tables
\item Marginal effects
\item Confidence intervals
\item Plots
\end{itemize}

\bigskip

\textbf{Required Output 1 — Marginal Effects}

Compute:

\begin{enumerate}
\item Effect of 1\% increase in immigrant stock
\item Effect of 1 additional migrant per 1,000 population
\end{enumerate}

Report:

\begin{itemize}
\item 95\% confidence intervals
\item Standard-deviation units (if possible)
\end{itemize}

\end{tcolorbox}


\begin{tcolorbox}[
    enhanced jigsaw,
    breakable,
    title={AI Coding Agent Protocol — Deliverables and Logging},
    colback=gray!5,
    colframe=black,
]

\textbf{Required Files}

Create:

\begin{enumerate}
\item research\_design.md
\item replication\_code.<ext>
\item results/marginal\_effects.csv
\item results/regression\_tables.md
\item results/plots/
\item conclusion.md
\item analysis\_log.txt
\end{enumerate}

\bigskip

\textbf{Substantive Conclusion}

Choose exactly one:

\begin{enumerate}
\item[(a)] Support
\item[(b)] Lack of support
\item[(c)] Not testable
\end{enumerate}

Provide justification.

\bigskip

\textbf{Analysis Log Must Include}

\begin{itemize}
\item Software versions
\item Data steps
\item Row counts
\item Implementation decisions
\item Errors and convergence issues
\end{itemize}

\end{tcolorbox}

\newpage

\begin{tcolorbox}[
    enhanced jigsaw,
    breakable,
    title={AI Coding Agent Protocol — Execution Rules},
    colback=gray!5,
    colframe=black,
]

\textbf{Rules}

\begin{enumerate}
\item Use R, Python, or Stata
\item Script must run end-to-end
\item Document decisions
\item Do not tune results
\item Do not run analyses during Phase II
\item Report failed models
\item Document infeasible tests
\item Do not consult prior published results
\item Do not modify source data directories
\end{enumerate}

\bigskip

\textbf{Output Directory}

\begin{verbatim}
/address/to/output/directory/
\end{verbatim}

\end{tcolorbox}

\section{Permission Settings}
\label{sec:settings}

\subsection{Claude Code}

\begin{tcolorbox}[title={Project-Level Configuration for Claude Code}, colback=gray!5, colframe=black, boxrule=0.5pt]

This guide describes how to configure a \texttt{settings.json} file for a \textbf{single Claude Code project} that:
\begin{itemize}[leftmargin=1.5em]
    \item Allows common development operations (editing files, running scripts, creating directories) without manual approval.
    \item Blocks all web access (including WebSearch, WebFetch, \texttt{curl}, and \texttt{wget}).
\end{itemize}

\begin{lstlisting}[style=terminal]
cd /path/to/your/project
\end{lstlisting}

\begin{lstlisting}[style=terminal]
mkdir -p .claude
\end{lstlisting}

Open the file in a text editor:

\begin{lstlisting}[style=terminal]
nano .claude/settings.json
\end{lstlisting}

\begin{lstlisting}[style=terminal]
cat .claude/settings.json
\end{lstlisting}

\vspace{0.5em}

Place the following content in \texttt{.claude/settings.json}:

\begin{lstlisting}[style=jsonstyle, label={lst:permissions}]
{
  "permissions": {
    "defaultMode": "acceptEdits",
    "allow": [
      "Bash(*)",
      "Write(*)",
      "Edit(*)",
      "MultiEdit(*)",
      "Read(*)"
    ],
    "deny": [
      "WebSearch",
      "WebFetch",
      "Bash(curl:*)",
      "Bash(wget:*)",
      "Bash(fetch:*)",
      "Read(~/.ssh/**)",
      "Read(~/.aws/**)",
      "Read(~/.env)",
      "Read(~/.gnupg/**)",
      "Edit(~/.bashrc)",
      "Edit(~/.zshrc)"
    ]
  },
  "sandbox": {
    "enabled": true,
    "autoAllowBashIfSandboxed": true
  }
}
\end{lstlisting}

\end{tcolorbox}

\subsection{Codex}

\begin{tcolorbox}[
    enhanced jigsaw,
    breakable,
    title={Codex Sandbox Configuration (config.toml)},
    colback=gray!5,
    colframe=black,
    boxrule=0.5pt,
    left=4pt,
    right=4pt,
    top=4pt,
    bottom=4pt
]

\begin{lstlisting}[language=toml]
#########################################################################
# Codex sandboxed reproducibility profile
# - Confines execution to the workspace (current directory + subdirs)
# - Disables Codex web search
# - Allows network only for package installation (pip / CRAN)
#########################################################################

sandbox_mode = "workspace-write"
approval_policy = "untrusted"
web_search = "disabled"

[sandbox_workspace_write]
network_access = true
exclude_slash_tmp = true
exclude_tmpdir_env_var = true
\end{lstlisting}

\end{tcolorbox}

\newpage

\section{Extended Results}

\subsection{Outcome-level distributional fidelity}

\begin{figure*}[h]
  \centering
  \includegraphics[width=\textwidth]{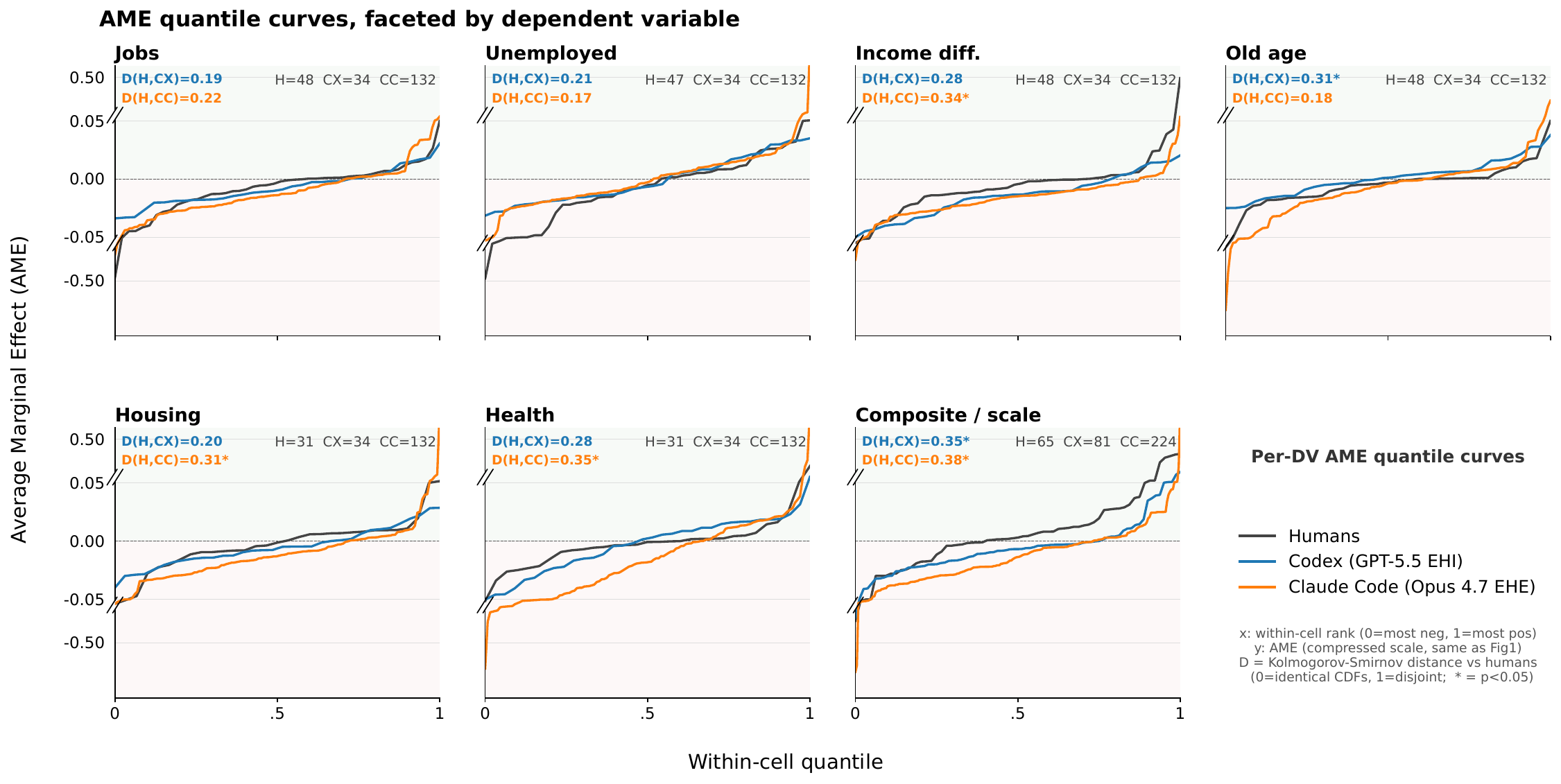}
  \caption{\textbf{Per-outcome AME distributions of AI agents track the human closely on most dependent variables but diverge on the composite scale.} Within-cell quantile curves of the average marginal effect (AME) of immigration on each policy outcome, comparing 73 human research teams (grey), 20 Codex (GPT-5.5 EHI) runs (blue), and 20 Claude Code (Opus 4.7 EHE) runs (orange). For every (DV, IV) cell, point estimates are ranked from the most negative ($x = 0$) to the most positive ($x = 1$); the curve traces the empirical CDF of AMEs that share each rank position. Sample sizes per panel are printed in the upper-right (H, humans; CX, Codex; CC, Claude Code). The $y$-axis uses the compressed scale of Fig.~1, with breaks at $|\text{AME}| = 0.05$ and $0.50$ to make small effects visible without truncating tails. $D$ values in the upper-left of each panel are two-sample Kolmogorov--Smirnov distances between the agent and the human distribution ($0 = $ identical CDFs, $1 = $ disjoint; asterisks mark $P < 0.05$). The two agents are statistically indistinguishable from humans on jobs, unemployment, and income difference, but Claude Code's distribution is significantly compressed on housing, health, and the composite scale ($D(\mathrm{H},\mathrm{CC}) = 0.21^{*},\, 0.35^{*},\, 0.38^{*}$), and Codex differs on old age and the composite ($D(\mathrm{H},\mathrm{CX}) = 0.31^{*},\, 0.35^{*}$).}
  \label{fig:fig5a}
\end{figure*}

\newpage

\subsection{Reproduction of Brady \& Finnigan (2014)}

\begin{figure*}[h]
  \centering
  \includegraphics[width=\textwidth]{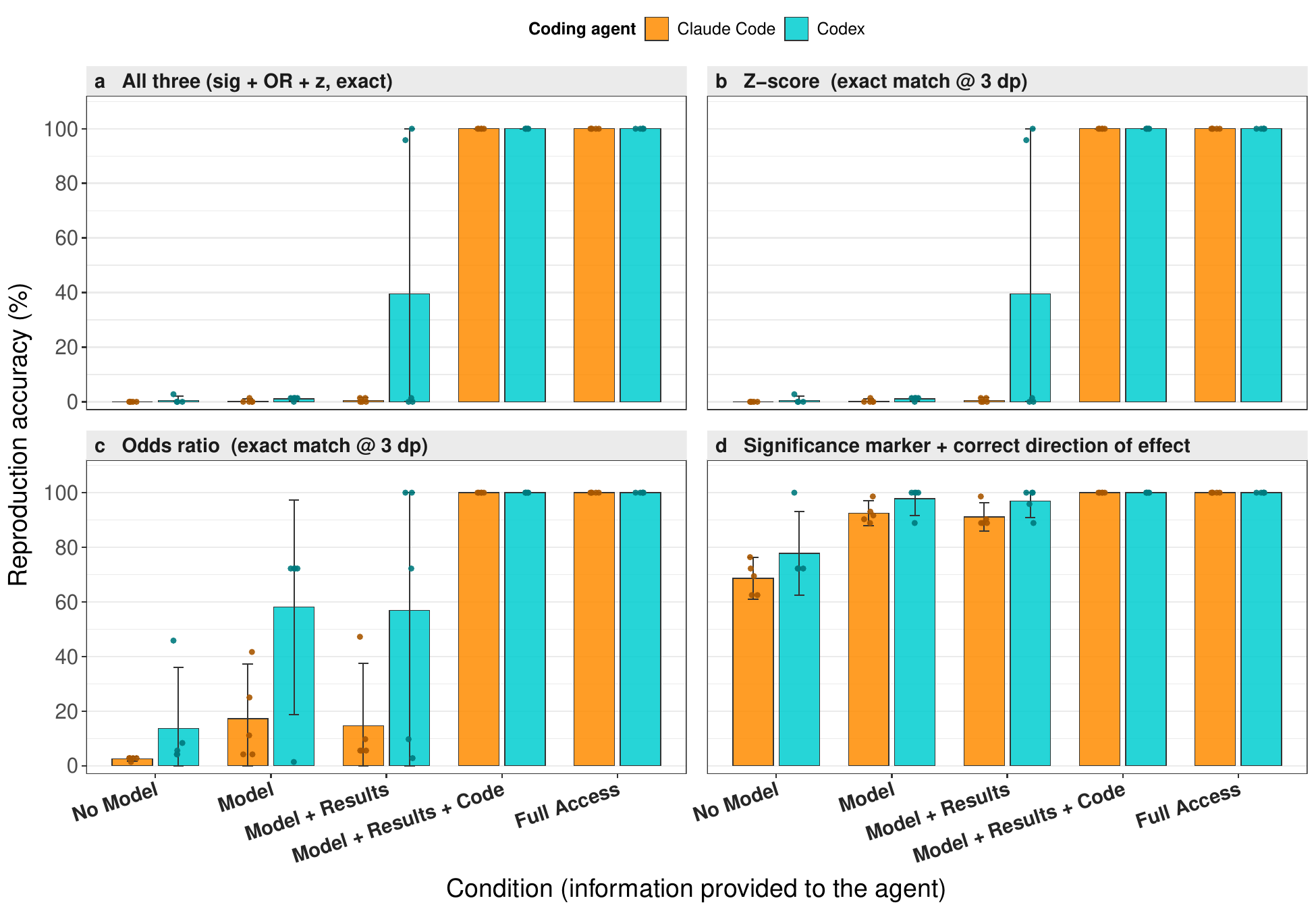}
  \caption{\textbf{Reproduction accuracy of two AI coding agents as a function of how much information from the original study is provided.} Each panel shows the per-cell reproduction accuracy (\%) of the Claude Code (orange) and OpenAI Codex (cyan) coding agents when asked to reproduce Table~1 of Brady \& Finnigan (2014) under five information conditions of increasing transparency: \textbf{No Model} (task description only), \textbf{Model} (statistical model/methods text), \textbf{Model + Results} (methods plus the published results table), \textbf{Model + Results + Code} (methods, results, and the original analysis code), and \textbf{Full Access} (the complete reproducibility package, including data). (\textbf{A})~Joint exact reproduction of all three reported quantities per cell---significance marker, odds ratio, and z-score (the last two matched to three decimal places). (\textbf{B})~Exact match of the z-score (3~d.p.). (\textbf{C})~Exact match of the odds ratio (3~d.p.). (\textbf{D})~Correct significance marker combined with the correct sign (direction) of the effect. Bars give the mean across $n=10$ independent runs per agent\,$\times$\,condition; dots show individual runs; error bars are 95\% confidence intervals (clipped at 0 and 100\%). Exact numerical reproduction (A--C) is essentially unattainable without access to the analysis code, jumping from $\leq\!1\%$ (Claude Code) or $\leq\!40\%$ (Codex, with very wide between-run variance) to $\sim\!100\%$ once code is supplied. In contrast, the qualitative pattern of significance and direction of effect (D) is recovered in $\sim\!70\text{--}80\%$ of cells from the methods text alone and exceeds 90\% once any model description is provided. Differences between the two agents are within run-to-run variability under every condition.}
  \label{fig:overview-accuracy}
\end{figure*}

We assessed whether two LLM-based coding agents could reproduce the 72 country-level coefficients reported in Tables~4 and 5 of Brady and Finnigan \cite{brady2014does} under five information conditions of increasing transparency, from the research question alone to full access to the authors' methods documentation and analysis code (SI Fig. \ref{fig:overview-accuracy}). Both agents converge to perfect reproduction once the original code is supplied: 100\% exact match on every metric under both \textit{Model + Results + Code} and \textit{Full Access}. Below this threshold, however, exact numerical reproduction is essentially unattainable. The joint exact match on the significance marker, odds ratio, and \textit{z}-score (each rounded to the paper's 3-decimal precision) stays below $1\%$ on average for Claude Code across all three partial-information conditions and is equally negligible for Codex when only the methods are supplied (1.1\%); Codex rises to a 39.4\% mean in the \textit{Model + Results} condition (SI Fig.~\ref{ffig:overview-accuracy}\textit{A}). The \textit{z}-score panel shows the same pattern (SI Fig.~\ref{fig:overview-accuracy}\textit{B}). Only the odds ratio alone partially survives this regime, rising to 17.2\% (Claude Code) and 58.1\% (Codex) when the model specification is provided (SI Fig.~\ref{fig:overview-accuracy}\textit{C}). Qualitative inference is far more robust: requiring only that the significance marker and the sign of the effect agree with the published value, accuracy reaches 68.6\%/77.8\% (Claude Code/Codex) from the methods section alone, climbs to 92.5\%/97.8\% once the regression specification is added, and exceeds 91\% across all model-aware conditions (SI Fig.~\ref{fig:overview-accuracy}\textit{D}).

Codex outperforms Claude Code in every partial-information condition, with the gap largest on the exact odds-ratio metric (Fig.~\ref{fig:overview-accuracy}\textit{C}: 58.1\% vs.\ 17.2\% under \textit{Model}; 56.9\% vs.\ 14.7\% under \textit{Model + Results}). Under \textit{Model + Results}, two of the five Codex runs achieve essentially perfect numerical reproduction ($\geq\!95.8\%$ on both the \textit{z}-score and the odds ratio), a third partially reproduces the odds ratio (72.2\%), and the remaining two cluster near zero, producing the long confidence intervals visible in panels~\textit{A--C}. A plausible mechanism is that some runs successfully recover the original Stata-equivalent estimation routine and rounding convention, while others adopt a different, but internally consistent, Python implementation whose third-decimal output diverges from the paper. Together, these results indicate that current coding agents can reliably reproduce the \emph{substantive} conclusions of an applied quantitative study from a methods section alone, but that bit-exact replication of the published numerical estimates remains effectively contingent on access to the original analysis code.

\newpage

\subsection{Explaining the Variability}
\label{sec:variance}

Having documented \emph{which} analytic decisions humans and AI
agents make, we now ask whether those decisions explain the
variability in their AME estimates and substantive conclusions. We extend the per-decision frequency table of Breznau et al \cite{breznau2022observing} (their SI Table S12) to all three of our groups; and the full extended table is in Appendix Table \ref{tab:s12_extended} (all 580 human/CX/CC cells filled).

A few quantitative contrasts stand out from the extended table. Of the 174 substantive decisions (Table~S12 minus 19 administrative identifiers and three PI-uncoded country rows), 26 are taken by at least half of all three groups' models, a consensus core. Beyond that the decision spaces diverge \emph{asymmetrically}: 32 decisions are present in human models but in $0\%$ of either agent's --- among them six Eastern-European country choices (Hungary, Latvia, Slovenia, Poland, Croatia, Russia), the M\textit{plus} factor-analytic measurement-model family, and several macro-control variants --- while only 4 decisions go the other way, present in $\geq\!50\%$ of both agents' models but $<\!10\%$ of humans': pure OLS (84\% CX, 60\% CC vs.\ 8\% humans), GDP per capita as a macro control (100\% / 100\% vs.\ 8\%), Belgium in the country sample (69\% / 58\% vs.\ 4\%), and treating the DV as categorical (57\% / 76\% vs.\ 4\%).

To translate these descriptive differences into an explanation of \emph{outcome} variance we apply the methodology of Breznau et al \cite{breznau2022observing}. For each group we fit a random-intercept linear mixed-effects model ($\texttt{lmer}$, REML) of the standardized AME on 15 decisions from their \emph{m07adj} block (DV indicators, measurement, sample, and model design), adopted as a fixed set rather than re-running their phase-wise AIC selection, with team or run as the random factor; variance reduction relative to the intercept-only baseline gives the between-team, within-team, and total explained shares. A multinomial logit on a four-predictor subset (\texttt{Stock}, \texttt{ChangeFlow}, \texttt{logit}, \texttt{twowayfe}) predicts team-level conclusions and reports deviance reduction. We drop two of the original Fig.~2 categories: \emph{Researcher Characteristics} (not measured for the AI agents, which also removes the competence-score term that distinguishes their preferred \emph{m13} from \emph{m07adj}), and \emph{Assigned Conditions} (both agents received an identical prompt, so there is no random task or deliberation assignment to vary). For stability of the variance components the human baseline uses the full 71-team CRI cohort ($n=1{,}253$ models); on the seed-42 subsample of 20 the between-team estimate inflates to $\sim$81\,\% via overfitting. The agent rows use the 20 runs each fixed by the experimental design.

\begin{figure}[t]
  \centering
  \includegraphics[width=\linewidth]{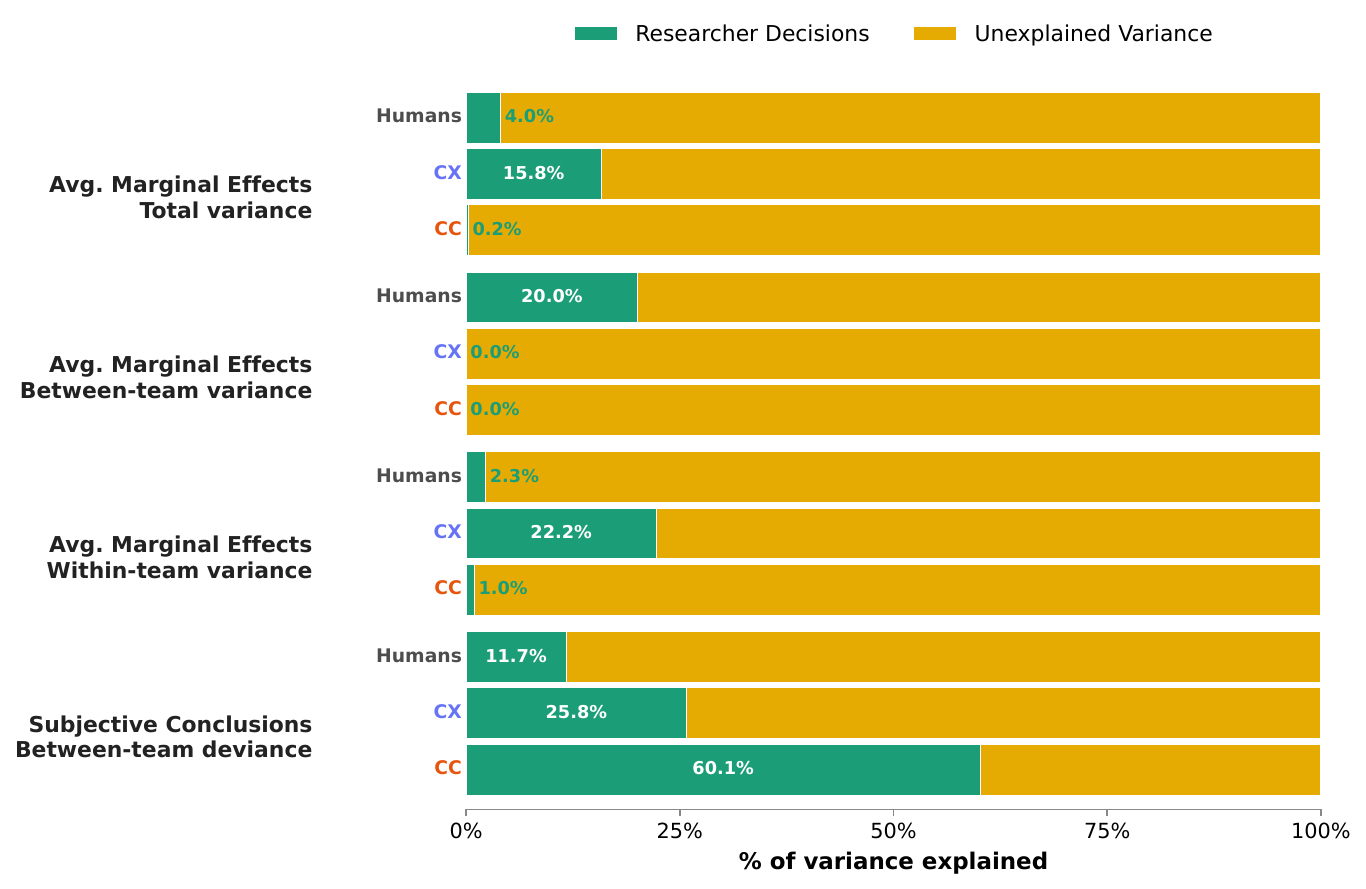}
  \caption{\textbf{Variance decomposition of standardized AMEs and subjective conclusions.} Bars show the percentage of variance (top three rows) or deviance (bottom row) explained by 15 researcher decisions --- Breznau et al.'s \emph{m07adj} block (DV choice, measurement, sample, model design), adopted as-is rather than re-running their phase-wise AIC selection; the researcher-aspect term that distinguishes their preferred \emph{m13} is omitted. The remainder is unexplained. (\textit{Top three rows}) AME variance decomposed with random-intercept mixed-effects regression (\texttt{lmer}, REML) using team or run as the random factor: \emph{between-team} compares team-level random-intercept variances, \emph{within-team} compares residual variances. (\textit{Bottom row}) Team-level conclusions (support / reject / not-testable / mixed) are predicted with a multinomial logit on a reduced 4-predictor set (\texttt{Stock}, \texttt{ChangeFlow}, \texttt{logit}, \texttt{twowayfe}); reductions are deviance-based against the intercept-only model. \emph{Sample sizes:} for stability of the variance components, the human baseline is computed on the full 73-team CRI cohort (71 teams with valid models, $n=1{,}253$ models), not the seed-42 subsample of 20; on the 20-team subsample the between-team estimate is unstable and inflates to $\sim$81\,\%.}
  \label{fig:variance_decomp}
\end{figure}

\paragraph{Codex's per-run AMEs are $10\times$ more decision-driven than humans'.}
For humans, the 15 decisions explain 4\% of total AME variance, 20\% between teams, 2\% within team, and 12\% of conclusion deviance. For Codex, the same 15 decisions explain 22\% of within-run AME variance and 16\% of total AME variance --- about $10\times$ the within-team value for humans (2\%) and $22\times$ the within-run value for Claude Code (1\%; Fig.~\ref{fig:variance_decomp}, middle two rows of the CX group). In other words, when Codex switches the estimator or the DV across the $\sim$18 models within a single run, those switches translate predictably into AME shifts. The implication is that Codex behaves comparatively deterministically: a single methodological choice has real leverage on its numerical estimates, leaving little of the kind of idiosyncratic per-model noise that ref. \cite{breznau2022observing} found dominated the human data. For an analyst inspecting a Codex spec curve, this means the spread of estimates can largely be \emph{traced back} to the decisions taken.

\paragraph{Claude Code's individual AMEs are unexplained but the hypothesis verdict is tightly decision-determined.}
For Claude Code, the 15 decisions explain only 0.2\% of total AME variance and 1\% within-run, yet individual AME estimates do not track which of those decisions are taken. Subjective-conclusion deviance, however, shows the inverse ordering: 60\% explained for CC, vs.\ 26\% for CX and 12\% for humans (Fig.~\ref{fig:variance_decomp}, bottom row). Claude Code's
eventual stance on the hypothesis is therefore tightly tied to its decisions even when its model-by-model numerical estimates are not. The split is consequential: in CC most of what drives any individual estimate lies \emph{below} the decision-level granularity (random seeds, data-prep idiosyncrasies, implementation details), but the run-level conclusion that summarises hundreds of those estimates is highly predictable from the decisions. Outputs that are noisy at the model level can nevertheless feed a narrative-level claim that is sharply decision-determined, a pattern worth flagging for any downstream user who reads only the executive summary of an agent's analysis.

\clearpage
\newpage

\subsection{Model Specification Coding and Distribution}

\input{Table_S12_extended.tex}

\newpage

\begin{figure*}[t]
  \centering
  \includegraphics[width=\linewidth]{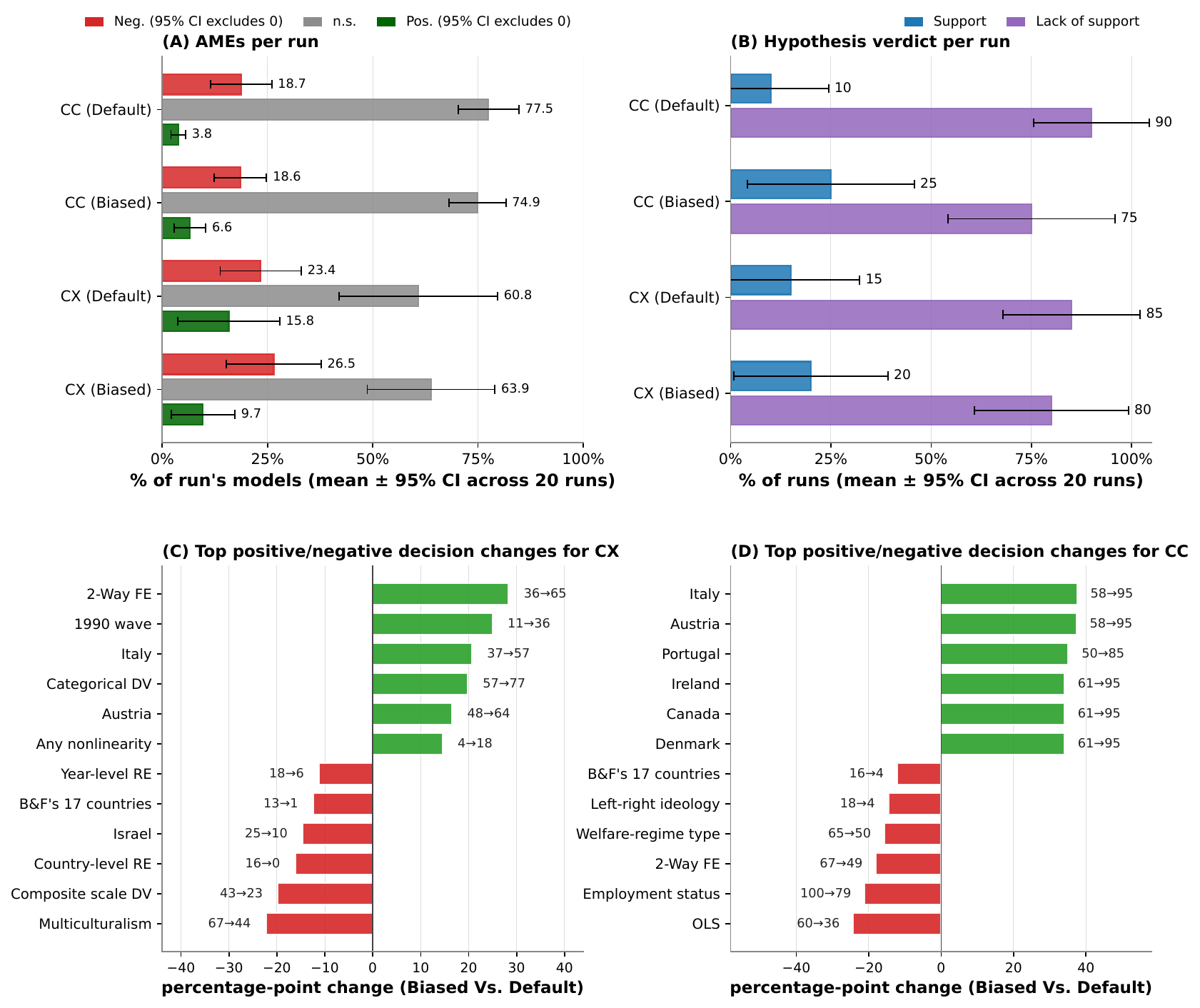}
  \caption{\textbf{Default Vs. Biased condition for Claude Code (CC) and Codex (CX).} (\textit{A})~Per-model AME conclusion mix
    from each run's 95\% CI relative to zero, partitioned into negative-significant ({\color[HTML]{D62728}red}), not significant (grey), and positive-significant ({\color[HTML]{006400}green}). Each row's three bars are the mean across the 20 runs of within-run percentages; horizontal whiskers are 95\% $t$-CIs of those means. (\textit{B})~Hypothesis-verdict mix per run, hand-coded from each run's as either ``support'' or ``lack of support''; bars and whiskers as in \textit{A}. (\textit{C}, \textit{D})~Top six positive and top six negative changes in adoption rate between the Default and Biased conditions for the 193 decisions, separately for Codex (\textit{C}) and Claude Code (\textit{D}). None of the per-metric Default--Biased differences is statistically separable at the 95\% level.}
  \label{fig:bias_comparison}
\end{figure*}

The four panels of Fig.~\ref{fig:bias_comparison} compare CC and CX between Default and Biased conditions, and none of the per-metric differences is statistically separable: the $95\%$ CIs of the Biased mean overlap the Default mean's CI in every panel and every metric. Models per run are $54.8$ vs $43.3$ for CC ([42.9, 66.7] vs [31.4, 55.2]; CIs overlap) and $17.9$ vs $14.1$ for CX ([11.5, 24.4] vs [12.1, 16.1]; CIs overlap), lower on average in the Biased runs of both agents. Pre-analysis-plan and replication-code lengths are also lower on average for CC under Biased ($1{,}128$ vs $972$ words; $859$ vs $789$ lines) and essentially the same for CX ($700$ vs $715$ words; $486$ vs $453$ lines), with overlapping CIs throughout. The per-model conclusion mix differs in opposite directions on the positive-significant share (CC $3.8\%$ vs $6.6\%$; CX $15.8\%$ vs $9.7\%$); hand-coded hypothesis verdicts are slightly higher in the Biased runs of both agents (CC $2/20$ vs $5/20$; CX $3/20$ vs $4/20$). All of these pairwise CIs again overlap. Beneath the aggregate patterns, the two agents' Biased$-$Default deltas across the 193 SI Table~S12 decisions are uncorrelated ($r = -0.03$): CC differs from its Default on $109$ of $193$ decisions ($55$ by $\geq\!10$ pp, $24$ by $\geq\!25$ pp; mean signed delta $+5.2$ pp), and CX differs on $68$ decisions ($16$ by $\geq\!10$ pp, $2$ by $\geq\!25$ pp; mean delta $+0.3$ pp). The asymmetries described below should therefore be read as descriptive patterns at $N=20$ runs per condition, not as effects detectable at conventional significance thresholds.

\newpage

\end{document}

%% file: Table_S12_extended.tex
\begingroup
\setlength{\tabcolsep}{4pt}
\renewcommand{\arraystretch}{1.10}
\scriptsize
\rowcolors{1}{gray!12}{white}
\begin{longtable}{@{}p{2.4cm}p{7.6cm}rrr@{}}
\hiderowcolors
\caption{Decision-frequency table extending Breznau et al.\ (2022) SI Table S12 to AI agents. Variable definitions are taken verbatim from S12 (with minor copy-edits for length). ``Humans'' are the 20 randomly sampled teams (seed $=42$) used in Fig.~1\textit{A}; ``CX'' are 20 Codex runs; ``CC'' are 20 Claude Code runs. Cells show the percentage of models in which the variable is coded as 1 / non-zero / present. Agent flags are derived by parsing each run's \texttt{marginal\_effects.csv}, \texttt{replication\_code.py}, and \texttt{conclusion.md} with pattern matchers tuned to the variable definitions; they should be read as automated approximations rather than exact recodes. Rows sorted by humans \% (desc.).}
\label{tab:s12_extended}\\
\toprule
\textbf{Variable} & \textbf{Definition} & \textbf{Humans} (\%) & \textbf{CX} (\%) & \textbf{CC} (\%) \\
\midrule
\showrowcolors
\endfirsthead
\hiderowcolors
\multicolumn{5}{l}{\emph{(continued from previous page)}}\\
\toprule
\textbf{Variable} & \textbf{Definition} & \textbf{Humans} (\%) & \textbf{CX} (\%) & \textbf{CC} (\%) \\
\midrule
\showrowcolors
\endhead
\hiderowcolors
\midrule
\multicolumn{5}{r}{\emph{(continued on next page)}}\\
\endfoot
\hiderowcolors
\bottomrule
\endlastfoot
\texttt{u\_teamid} & Random team number assignment except team 0, which refers to the Brady and Finnigan study. These specifications are excluded from the analysis but left in here for comparison. & 100 & 100 & 100 \\
\texttt{main\_IV\_type} & Test variable type for the hypothesis that immigration undermines social policy support: "stock" (\% foreign-born), "flow" (change in \%, net migration or change in stock), or "change in flow". & 100 & 100 & 100 \\
\texttt{count} & A counter to return results to their original order. & 100 & 100 & 100 \\
\texttt{num\_countries} & Number of countries in the model sample. & 100 & 100 & 100 \\
\texttt{inv\_weight} & The number of models per team, must be divided into 1 to use for weighting. & 100 & 100 & 100 \\
\texttt{main\_IV\_effect} & Total, within, or between effect. For non-multilevel models, always total. A within-effect of stock is "Flow per wave". & 100 & 100 & 100 \\
\texttt{main\_IV\_time} & The time period the team used to measure flow of immigrants (1-year, 5-year, etc.). PIs rescaled to a 1-year equivalent for comparability; this refers to the original metric. & 100 & 100 & 100 \\
\texttt{main\_IV\_measurement} & Measuring what type of immigrants. "Emigration" is coded as "Immigrant, foreign-born". & 100 & 100 & 100 \\
\texttt{main\_IV\_source\_file} & Name of the source file used. & 100 & 100 & 100 \\
\texttt{main\_IV\_source} & The data source; many teams imputed some countries using other sources, coded only as the primary source. (Deprecated.) & 100 & 100 & 100 \\
\texttt{package} & Software package, character categories. & 100 & 100 & 100 \\
\texttt{DV} & Dependent variable used; single questions labeled "Jobs" etc.; scale variables start with "Scale\_" followed by the number of items. & 100 & 100 & 100 \\
\texttt{z} & Z-statistic or equivalent (T-value). & 100 & 100 & 100 \\
\texttt{error} & The absolute deviation of the high 95\% CI from the margin. & 100 & 100 & 100 \\
\texttt{upper} & Upper confidence boundary at 95\% CI. & 100 & 100 & 100 \\
\texttt{lower} & Lower confidence boundary at 95\% CI. & 100 & 100 & 100 \\
\texttt{AME} & Average marginal effect as produced by team's provided code; or added by PIs to produce when not present. & 100 & 100 & 100 \\
\texttt{p} & p-value or equivalent confidence interval relative to zero (e.g. for Bayes estimation). & 100 & 100 & 100 \\
\texttt{id} & Team number plus model number counted in order within teams. & 100 & 100 & 100 \\
\texttt{sex\_iv} & Sex / gender of respondent. & 97.49 & 100 & 100 \\
\texttt{age\_iv} & Age as a continuous variable. & 97.49 & 100 & 100 \\
\texttt{switzerland} & Country included in sample. & 96.24 & 95.82 & 61.22 \\
\texttt{france} & Country included in sample. & 96.24 & 95.82 & 61.22 \\
\texttt{norway} & Country included in sample. & 96.24 & 95.82 & 61.22 \\
\texttt{spain} & Country included in sample. & 96.24 & 95.82 & 61.22 \\
\texttt{sweden} & Country included in sample. & 96.24 & 95.82 & 61.22 \\
\texttt{w2006} & Includes data from ISSP 2006 wave. & 94.98 & 100 & 99.82 \\
\texttt{australia} & Country included in sample. & 93.73 & 95.82 & 91.24 \\
\texttt{usa} & Country included in sample. & 92.48 & 100 & 91.24 \\
\texttt{listwise} & Listwise deletion: cases are dropped if any relevant variable is missing for that observation. & 91.22 & 100 & 100 \\
\texttt{germany} & Country included in sample. & 90.60 & 95.82 & 61.22 \\
\texttt{employed\_iv} & Employed, or a categorical variable with self / public / full / part etc. & 89.97 & 61.84 & 100 \\
\texttt{education\_iv} & Any measure of educational attainment or years (rough; finer-grained coding could be considered). & 89.97 & 100 & 100 \\
\texttt{new\_zealand} & Country included in sample. & 89.34 & 95.82 & 61.22 \\
\texttt{w1996} & Includes data from ISSP 1996 wave. & 88.71 & 100 & 98.72 \\
\texttt{great\_britain} & Country included in sample. & 84.95 & 95.82 & 61.41 \\
\texttt{age2\_iv} & Age-squared, or a categorical break-down (a non-linear age function). & 84.33 & 100 & 94.53 \\
\texttt{japan} & Country included in sample. & 79.94 & 95.82 & 61.22 \\
\texttt{w2016} & Includes data from ISSP 2016 wave. & 78.37 & 91.09 & 80.11 \\
\texttt{canada} & Country included in sample. & 68.34 & 95.82 & 61.22 \\
\texttt{dichotomize} & Dependent variable is dichotomized. & 63.95 & 0 & 1.09 \\
\texttt{stata} & Stata software employed (dummy for package). & 63.64 & 0 & 0 \\
\texttt{logit} & Logistic regression; fits "S"-shaped logistic curve to a 0/1 DV. Includes multilevel logistic. & 62.07 & 0 & 17.70 \\
\texttt{ireland} & Country included in sample. & 58.93 & 95.82 & 61.22 \\
\texttt{finland} & Country included in sample. & 57.68 & 77.16 & 61.22 \\
\texttt{Stock} & Dichotomous indicator for main\_IV\_type. & 56.74 & 53.20 & 56.75 \\
\texttt{income\_iv} & Income. & 53.29 & 91.64 & 88.78 \\
\texttt{mlm\_any} & Any multilevel model: =1 if mlm\_re, mlm\_fe, and/or hybrid\_mlm =1. & 52.04 & 0 & 7.66 \\
\texttt{unbalpanel} & Unbalanced time-series; includes different numbers of countries per wave. & 50.47 & 100 & 100 \\
\texttt{twowayfe} & Two-way fixed-effects (2WFE). Contains dummy variables for country and year regardless of estimation strategy. The PIs follow the Brady-Finnigan nomenclature. & 48.59 & 36.21 & 66.61 \\
\texttt{mlm\_re} & Random-effects multilevel model: random intercepts and fixed coefficients (an "RE model" in econometrics). & 46.39 & 0 & 3.65 \\
\texttt{Hmixed} & Two separate, internally consistent conclusions about stock and flow leading to mixed-result claims. & 46.39 & 29.53 & 48.63 \\
\texttt{level\_country} & Unspecified modelling of country level, can include random-effects or dummies. & 45.14 & 16.16 & 0 \\
\texttt{denmark} & Country included in sample. & 44.83 & 77.16 & 61.22 \\
\texttt{Flow} & Dichotomous indicator for main\_IV\_type. & 41.38 & 46.80 & 43.25 \\
\texttt{emplrate\_ivC} & Employment rate (usually of those in the labor force). & 40.13 & 91.36 & 35.86 \\
\texttt{Hreject\_stock} & Hypothesis rejected specifically for stock (see above). & 36.99 & 0 & 45.80 \\
\texttt{socx\_ivC} & Social Expenditures \% of GDP ("SOCX"). & 34.80 & 100 & 85.95 \\
\texttt{Unemp} & Single question on government provision of unemployment protection is the DV, or part of the scale if Scale=1. & 32.92 & 9.47 & 12.96 \\
\texttt{OldAge} & Single question on government provision of old-age care is the DV, or part of the scale if Scale=1. & 32.92 & 9.47 & 12.77 \\
\texttt{IncDiff} & Single question on government reduction of income differences is the DV, or part of the scale if Scale=1. & 32.92 & 9.47 & 12.77 \\
\texttt{main\_IV\_as\_control} & If the other main IV is in the same model: 0=no, 1=yes. Within/between models =1 only if both stock and flow are entered as separate variables. & 31.97 & 0 & 0 \\
\texttt{portugal} & Country included in sample. & 31.97 & 46.80 & 50.00 \\
\texttt{Hsupport\_net} & Hypothesis supported specifically for the flow / net-migration test variable. & 31.35 & 65.18 & 27.92 \\
\texttt{Jobs} & Single question on government provision of jobs is the DV, or part of the scale if Scale=1. & 30.41 & 9.47 & 12.77 \\
\texttt{netherlands} & Country included in sample. & 28.21 & 77.16 & 61.22 \\
\texttt{r} & R software employed (dummy for package). & 27.59 & 0 & 17.06 \\
\texttt{eeurope} & Includes at least 3 Eastern European countries. & 27.59 & 3.90 & 8.94 \\
\texttt{hungary} & Country included in sample. & 27.59 & 0 & 0 \\
\texttt{latvia} & Country included in sample. & 27.59 & 0 & 0 \\
\texttt{Hreject\_net} & Hypothesis rejected specifically for flow / net-migration. & 26.33 & 0 & 11.13 \\
\texttt{slovenia} & Country included in sample. & 26.33 & 0 & 0 \\
\texttt{Hsupport\_stock} & Hypothesis supported specifically for the stock test variable (only listed when researchers report stock/flow conclusions separately). & 26.33 & 95.82 & 43.80 \\
\texttt{House} & Single question on government provision of housing is the DV, or part of the scale if Scale=1. & 25.71 & 9.47 & 12.77 \\
\texttt{Health} & Single question on government provision of health care is the DV, or part of the scale if Scale=1. & 25.71 & 9.47 & 12.77 \\
\texttt{Hreject} & Researchers conclude the hypothesis is rejected; inconclusive support is also counted as rejection. & 25.08 & 19.50 & 15.78 \\
\texttt{level\_cyear} & Unspecified modelling of country-year level, can include random-effects or dummy variables in a multilevel model. & 23.82 & 0 & 0.18 \\
\texttt{mmodel} & Measurement model: uses scaling, factor analysis or item-response to test/generate a latent DV. Always with a linear estimator. & 23.82 & 0 & 0 \\
\texttt{czechia} & Country included in sample. & 23.82 & 0 & 5.47 \\
\texttt{poland} & Country included in sample. & 22.57 & 0 & 0 \\
\texttt{ml\_glm} & Maximum likelihood: ML or any other iterative version that is not OLS, Bayes or Logit (e.g., GLM, MWFE). & 22.57 & 0 & 1.09 \\
\texttt{Scale} & A multi-item scale was constructed and used as the DV; the questions used are indicated by the previous 6 variables. & 20.38 & 43.18 & 23.18 \\
\texttt{allavailable} & \textgreater{}21 countries; all available or mostly all. & 20.06 & 3.90 & 8.94 \\
\texttt{Hsupport} & Researchers conclude immigration undermines social-policy preferences and the team's evidence supports it (subjective; team prerogative). & 19.44 & 66.57 & 60.86 \\
\texttt{croatia} & Country included in sample. & 18.81 & 0 & 0 \\
\texttt{israel} & Country included in sample. & 18.81 & 24.51 & 14.69 \\
\texttt{korea} & Country included in sample. & 18.81 & 3.90 & 9.12 \\
\texttt{year\_dummies\_only} & If not 2WFE: includes a year dummy for each year (also includes dummies within an MLM but not RE intercepts). & 15.67 & 0 & 1.28 \\
\texttt{orig13} & Identical to the original 13 countries used in Brady \& Finnigan's two-way fixed-effects models ("13 richest democracies"). & 15.67 & 13.09 & 7.03 \\
\texttt{russia} & Country included in sample. & 15.05 & 0 & 0 \\
\texttt{leftright\_iv} & Left-right subjective political ideology, or actual reported party vote coded into left/right. & 13.79 & 0 & 18.16 \\
\texttt{level\_year} & Unspecified modelling of year level, can include random-effects or dummies. Refers technically to survey wave. & 12.54 & 17.55 & 2.55 \\
\texttt{socialistdummy\_ivC} & Former state-socialist societies = 1, others = 0. & 11.29 & 51.81 & 50.36 \\
\texttt{italy} & Country included in sample. & 10.66 & 36.77 & 57.76 \\
\texttt{w1990} & Includes data from ISSP 1990 wave. & 10.66 & 10.86 & 30.75 \\
\texttt{fract\_ivC} & Ethnic fractionalization / Herfindahl index (e.g., from UN stock-by-origin data, Alesina). & 10.03 & 41.50 & 40.88 \\
\texttt{anynonlin} & Any nonlinearity used; =1 if any of the above interactions =1, plus a few cases with interactions not in the list (e.g., team-98 immigration x party voting; one squared-DV in team 29). & 9.40 & 3.90 & 2.37 \\
\texttt{Hnotest} & Researchers conclude the hypothesis is not testable, or the evidence is inconclusive to support or reject. & 9.09 & 4.46 & 3.47 \\
\texttt{mplus} & Mplus software employed (package dummy). & 8.78 & 0 & 0 \\
\texttt{unemprate\_ivC} & Unemployment rate of those in the labor force (usually means registered unemployed). & 8.46 & 0 & 0 \\
\texttt{mcp\_ivC} & Multiculturalism Policy Index, MIPEX, or IMPIC immigration policies index. & 8.15 & 66.57 & 39.60 \\
\texttt{ols} & Ordinary least squares estimator. & 7.84 & 84.40 & 59.85 \\
\texttt{bulgaria} & Country included in sample. & 7.52 & 0 & 0 \\
\texttt{chile} & Country included in sample. & 7.52 & 0 & 0 \\
\texttt{hybrid\_mlm} & Includes both random-effects and fixed-effects components. & 7.52 & 0 & 4.11 \\
\texttt{gdp\_ivC} & GDP per capita. & 7.52 & 100 & 100 \\
\texttt{south\_africa} & Country included in sample. & 7.52 & 0 & 0 \\
\texttt{cyprus} & Country included in sample. & 7.52 & 0 & 0 \\
\texttt{household\_iv} & Household composition (unspecified). & 6.90 & 0.56 & 17.52 \\
\texttt{mlm\_fe} & Fixed-effects multilevel model: random intercepts so country-level variables are mean-centered within country; explains within-country changes only. & 5.64 & 0 & 4.01 \\
\texttt{Hnotest\_net} & Hypothesis not testable specifically for flow / net-migration. & 5.64 & 0 & 0 \\
\texttt{fbXleftright} & Interaction (indicated by "X"). & 5.02 & 0 & 0 \\
\texttt{bayes} & Bayesian estimator (MCMC etc.) fitting posterior probabilities based on prior distributions for more 'consistent' level-2 estimates. & 5.02 & 0 & 0 \\
\texttt{w1985} & Includes data from ISSP 1985 wave. & 5.02 & 3.90 & 8.76 \\
\texttt{cluster\_any} & Any kind of clustering command added by the researcher (excludes a multilevel model's implicit clustering). & 4.70 & 85.79 & 35.95 \\
\texttt{slovakia} & Country included in sample. & 4.39 & 0 & 0 \\
\texttt{belgium} & Country included in sample. & 4.39 & 68.80 & 57.76 \\
\texttt{orig17} & Identical to the 17 countries used by Brady \& Finnigan in their MLM random-effects models. & 4.39 & 13.09 & 16.24 \\
\texttt{weights} & Any survey weights applied. & 4.08 & 3.90 & 4.56 \\
\texttt{mlogit} & Multinomial logistic estimator. Includes multilevel ordered logit or probits. & 3.76 & 0 & 0 \\
\texttt{netXinc} & Interaction (indicated by "X"). & 3.76 & 0 & 0 \\
\texttt{categorical} & Dependent variable has more than 2 categories. & 3.76 & 56.82 & 76.09 \\
\texttt{iceland} & Country included in sample. & 3.76 & 39.28 & 39.05 \\
\texttt{L2boots} & Robust SE or bootstrapped level-2 analysis (jackknife, sandwich robust, or fe-robust in Stata's xtreg). & 3.76 & 0 & 2.55 \\
\texttt{married\_iv} & Marital status. & 3.45 & 0 & 0 \\
\texttt{decomm\_ivC} & Some measure of replacement rates (Scruggs / CWED). & 3.13 & 0 & 0 \\
\texttt{conservatism\_ivC} & Conservative (left-vs-right) government political-ideology index (e.g., Schmidt index); includes vote-share measures. & 3.13 & 0 & 5.47 \\
\texttt{ologit} & Ordered logistic / probabilistic estimator (probit). Includes item-response, ordered-logit and probit models. & 2.51 & 0 & 5.66 \\
\texttt{lpm} & Linear probability model estimation. DV coded 0/1 but linear model used. & 2.51 & 10.03 & 6.66 \\
\texttt{socult\_ivC} & Socio-cultural proximity scale using country of origin for immigrants. & 2.51 & 0 & 0 \\
\texttt{pseudo\_pnl} & Constructed a pseudo-panel of individual-level groups. & 1.88 & 0 & 0 \\
\texttt{taiwan} & Country included in sample. & 1.88 & 0 & 0 \\
\texttt{lithuania} & Country included in sample. & 1.88 & 0 & 0 \\
\texttt{ChangeFlow} & Dichotomous indicator for main\_IV\_type. & 1.88 & 2.23 & 4.74 \\
\texttt{india} & Country included in sample. & 1.88 & 0 & 0 \\
\texttt{turkey} & Country included in sample. & 1.88 & 0 & 0 \\
\texttt{austria} & Country included in sample. & 1.88 & 47.63 & 57.94 \\
\texttt{ginin\_ivC} & Gini (not enough cases of pre-tax Gini to differentiate; also includes one case of top-income concentration from WID). & 1.88 & 0.56 & 78.92 \\
\texttt{multimpute} & Pairwise information or imputation employed (e.g. FIML or multiple imputation). & 1.25 & 0 & 0 \\
\texttt{year\_as\_count} & Year added as a continuous variable; =1 if year is continuous and \textgreater{}2 waves are included. & 1.25 & 0 & 0 \\
\texttt{fbXeduc} & Interaction (indicated by "X"). & 0.31 & 0 & 2.37 \\
\texttt{netXeduc} & Interaction (indicated by "X"). & 0.31 & 0 & 0 \\
\texttt{fbXnet} & Interaction (indicated by "X") between foreign-born stock and net migration. & 0 & 0 & 0 \\
\texttt{netXcons} & Interaction: net migration x conservatism index. & 0 & 0 & 0 \\
\texttt{netXage} & Interaction (indicated by "X"). & 0 & 0 & 0 \\
\texttt{efficacy\_iv} & Political efficacy (believes he/she can influence government). & 0 & 0 & 0 \\
\texttt{netXsex} & Interaction (indicated by "X"). & 0 & 0 & 0 \\
\texttt{netXunemp} & Interaction (indicated by "X"). & 0 & 0 & 0 \\
\texttt{fractXfb} & Interaction (indicated by "X"). & 0 & 0 & 0 \\
\texttt{fbXage} & Interaction (indicated by "X"). & 0 & 0 & 0 \\
\texttt{fbXsex} & Interaction (indicated by "X"). & 0 & 0 & 0 \\
\texttt{squared\_imm} & A quadratic form for one or both immigration variables. & 0 & 3.90 & 0 \\
\texttt{Hnotest\_stock} & Hypothesis not testable specifically for stock (see above). & 0 & 0 & 0 \\
\texttt{fbXunemp} & Interaction (indicated by "X"). & 0 & 0 & 0 \\
\texttt{fbXgini} & Interaction (indicated by "X"). & 0 & 0 & 0 \\
\texttt{fbXurban} & Interaction (indicated by "X"). & 0 & 0 & 0 \\
\texttt{fbXinc} & Interaction (indicated by "X"). & 0 & 0 & 0 \\
\texttt{philippines} & Country included in sample. & 0 & 0 & 0 \\
\texttt{trust\_iv} & Political trust. & 0 & 0 & 0 \\
\texttt{mlwin} & MLwiN software (package dummy). & 0 & 0 & 0 \\
\texttt{upol\_iv} & Subjective interest in politics. & 0 & 0 & 0 \\
\texttt{taxes\_iv} & Subjective attitude that government should tax more / less. & 0 & 0 & 0 \\
\texttt{uruguay} & Country included in sample. & 0 & 0 & 0 \\
\texttt{emigration\_ivC} & Gross or net out-migration ('flow'). & 0 & 0 & 0 \\
\texttt{unchange\_ivC} & Annual change in unemployment rate. & 0 & 0 & 0 \\
\texttt{poverty\_ivC} & Poverty (e.g. 50\% of median). & 0 & 0 & 0 \\
\texttt{fbunemprate\_ivC} & Foreign-born unemployment rate. & 0 & 0 & 0 \\
\texttt{fbunempchange\_ivC} & Change in foreign-born unemployment rate. & 0 & 0 & 0 \\
\texttt{fbeducrate\_ivC} & Foreign-born education rate. & 0 & 0 & 0 \\
\texttt{fbeducratechange\_ivC} & Change in foreign-born education rate. & 0 & 0 & 0 \\
\texttt{socxchg\_ivC} & Change in SOCX. & 0 & 0 & 0 \\
\texttt{gdpchange\_ivC} & Any change measure of GDP (1-yr / 5-yr, etc.). & 0 & 0 & 0 \\
\texttt{regime\_ivC} & Categorical welfare-state or institutional-regime type, not including a post-communist split. & 0 & 0 & 65.33 \\
\texttt{targeting\_ivC} & Benefits target groups (vs.\textbackslash\{\} universal). & 0 & 0 & 0 \\
\texttt{socx\_programspecific\_ivC} & Social spending decomposed into single program domains. & 0 & 0 & 0 \\
\texttt{subFB\_ivC} & Subjective foreign-born, country mean. & 0 & 0 & 0 \\
\texttt{spss} & SPSS software employed (dummy). & 0 & 0 & 0 \\
\texttt{antiimm\_ivC} & Aggregate measures of anti-immigrant attitudes / sentiment from other surveys (e.g., ISSP National Identity, ESS). & 0 & 0 & 5.66 \\
\texttt{pop\_ivC} & Population of country. & 0 & 0 & 0 \\
\texttt{occclass\_iv} & Occupational class. & 0 & 0 & 0 \\
\texttt{occstatus\_iv} & Occupational status. & 0 & 0 & 0 \\
\texttt{country\_dummies\_only} & If not 2WFE: includes a country dummy for each country (also includes dummies within an MLM but not RE intercepts). & 0 & 0 & 0 \\
\texttt{venezuela} & Country included in sample. & 0 & 0 & 0 \\
\texttt{reldenom\_iv} & Religious denomination. & 0 & 0 & 0 \\
\texttt{relattend\_iv} & Religious service attendance. & 0 & 0 & 16.70 \\
\texttt{publice\_iv} & Employed in the public sector. & 0 & 0 & 0 \\
\texttt{urban\_iv} & Urban / rural / suburban (unspecified). & 0 & 0 & 0 \\
\texttt{fb\_iv} & Foreign-born respondent in the ISSP. & 0 & 5.85 & 58.12 \\
\texttt{cuts\_iv} & Subjective attitude that government should make cuts. & 0 & 0 & 10.40 \\
\texttt{tradeunion10\_ivC} & 10-year change in trade-union share of employed. & 0 & 0 & 0 \\
\texttt{germany\_west} & Distinguished (not coded, not enough cases). & \textemdash & 0 & 0 \\
\texttt{germany\_east} & Distinguished (not coded, not enough cases). & \textemdash & 0 & 0 \\
\texttt{n\_ireland} & Distinguished (not coded, not enough cases). & \textemdash & 0 & 0 \\
\end{longtable}
\endgroup